\documentclass[pdflatex,sn-mathphys-num]{sn-jnl}

\usepackage{graphicx}%
\usepackage{multirow}%
\usepackage{amsmath,amssymb,amsfonts}%
\usepackage{amsthm}%
\usepackage{mathrsfs}%
\usepackage[title]{appendix}%
\usepackage{xcolor}%
\usepackage{textcomp}%
\usepackage{manyfoot}%
\usepackage{booktabs}%
\usepackage{algorithm}%
\usepackage{algorithmicx}%
\usepackage{algpseudocode}%
\usepackage{listings}%
\usepackage{array}%

\theoremstyle{thmstyleone}%
\theoremstyle{thmstyletwo}%

\theoremstyle{thmstylethree}%

\begin{document}

\title[Article Title]{Large scale cross-regional remote sensing flood monitoring framework for operative mapping and impact analysis}

\author[1]{\fnm{Ilya} \sur{Novikov}}\email{ilya.novikov2@skoltech.ru}
\author*[1]{\fnm{Svetlana} \sur{Illarionova}}\email{s.illarionova@skoltech.ru}
\author[2]{\fnm{Ruslan} \sur{Dzharkinov}}\email{ruslandzarkinov@gmail.com}
\author[1]{\fnm{Maria} \sur{Smirnova}}\email{mariya.smirnova2@skoltech.ru}
\author[3]{\fnm{Ayrat} \sur{Abdullin}}\email{g202203180@kfupm.edu.sa}
\author[4]{\fnm{Anna} \sur{Korotkova}}\email{annkornn@mail.ru}
\author[5]{\fnm{Mariia} \sur{Ulianova}}\email{ulyanova.ma@phystech.edu}
\author[1]{\fnm{Dmitrii} \sur{Shadrin}}\email{d.shadrin@skoltech.ru}
\author[1]{\fnm{Evgeny} \sur{Burnaev}}\email{e.burnaev@skoltech.ru}

\affil*[1]{\orgname{Skolkovo Institute of Science and Technology}, \orgaddress{\city{Moscow}, \postcode{121205}, \country{Russia}}}
\affil[2]{\orgname{Trofimuk Institute of Petroleum Geology and Geophysics SB RAS}, \orgaddress{\city{Novosibirsk}, \postcode{630090}, \country{Russia}}}
\affil[3]{\orgname{King Fahd University of Petroleum and Minerals}, \orgaddress{\city{Dhahran}, \postcode{31261}, \country{Saudi Arabia}}}
\affil[4]{\orgname{Tyumen Industrial University}, \orgaddress{\city{Tyumen}, \postcode{625000}, \country{Russia}}}
\affil[5]{\orgname{Huawei Russian Research Institute}, \orgaddress{\city{Moscow}, \country{Russia}}}

\abstract{Effective flood monitoring is critical for minimizing the impacts of flood disasters on populations and infrastructure. Yet reliable remote sensing across extensive and environmentally diverse regions remains challenging, as most segmentation algorithms lack the generalisation capacity required for large-scale application, while annotated flood data are scarce and unevenly distributed. This problem is particularly acute  for the Russian Federation, where climatic, hydrological and flood-generation regimes vary dramatically across the country and ground-based monitoring is limited in many areas.

This study presents an end-to-end multimodal framework for flood monitoring and damage assessment based on synthetic aperture radar data, multispectral imagery, and digital elevation models with their derivatives, forming a 21-channel input. Using a self-collected multimodal dataset covering seven Russian regions, two strategies for water surface detection under limited data conditions were compared: a supervised U-Net++ model and the self-supervised AnySat architecture pre-trained and fine-tuned for the segmentation task. Under the data conditions of this study, supervised learning proved more effective, with the U-Net++ multimodal configuration achieving a mean F1-score of $0.84 \pm 0.11$ under region-based cross-validation, while the AnySat-based approach offered greater stability and retains advantages for settings where larger unlabelled data or missing modalities at inference are expected.

The best flood area predictions were used to estimate flood impact in urban areas in terms of the area affected, material damage, casualties, and ecological and agricultural impact. The estimations were conducted following the official methodology of the Russian Ministry of Emergency Situations. Applied to the 2019 Tulun flood, the obtained results closely matched official assessments, except for material damage, due to the open-source databases usage. The results demonstrate the potential of deep learning and multimodal satellite data integration for scalable, reliable flood monitoring across diverse environmental and data-limited conditions.}


\maketitle

\maketitle

\newpage

\begin{figure*}[h]
    \centering
    \includegraphics[scale=0.052]{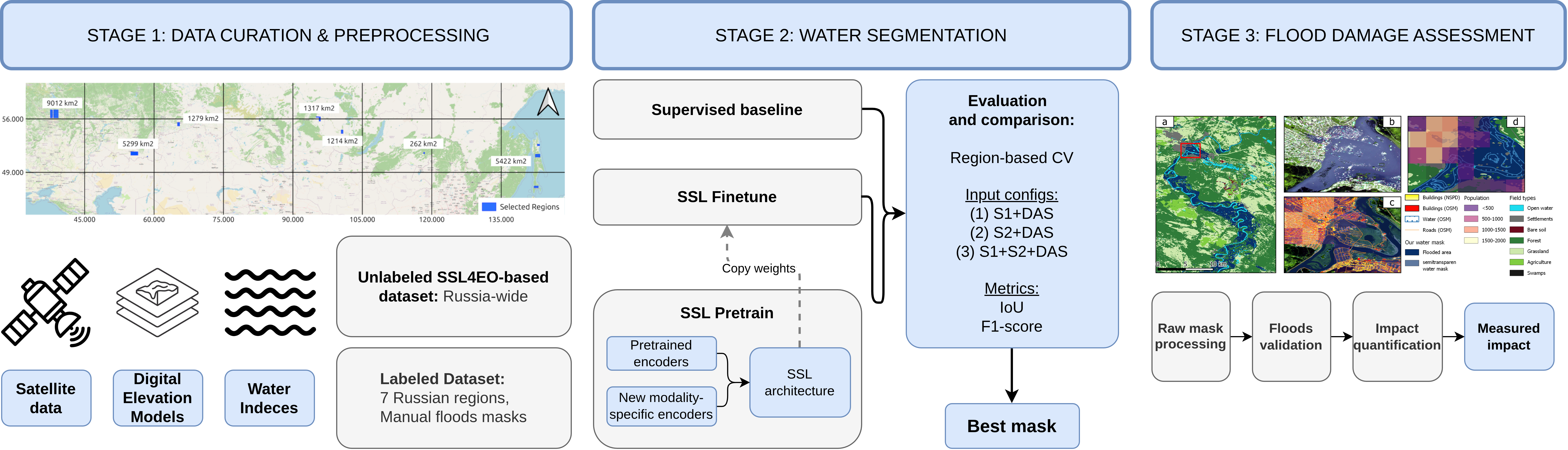}
    \caption{Graphical Abstract Descriptions: this visual summary illustrates the complete three-stage mutimodal pipeline of the proposed flood monitoring and damage assessment framework. Stage 1 (Data Curation and Preprocessing) shows how multimodal inputs are constructed: an unlabelled, Russia-wide corpus of 7,912 patches for self-supervised learning based on SSL4EO-S12 dataset~\cite{wang2023ssl4eo} and a manually labelled dataset of 1,259 patches across seven Russian regions, combining Sentinel-1, Sentinel-2, water indices, and Digital Elevation Models with derived Slope and Aspect (DAS). Stage 2 (Water Segmentation) depicts the core methodological comparison between a supervised U-Net++ baseline and a self-supervised AnySat model that is pre-trained and then fine-tuned with a decoder head. Both pathways are evaluated under region-based cross-validation across three modality configurations using IoU and F1-score, yielding the best water mask. Stage 3 (Flood Damage Assessment) illustrates step-by-step flood mask post processing for impact quantification. The color-coded left-to-right flow conveys the logical progression from data to model to real-world consequence and highlights the study's central message: multimodal, data-efficient deep learning can deliver scalable, reliable flood monitoring and actionable impact estimates across diverse regions.}
    \label{fig:graphical_abstract}
\end{figure*}

\newpage

Highlights

\begin{itemize}
    \item End-to-end multimodal framework enables cross-regional flood mapping and damage assessment across climatically diverse regions. 
    \item Considered 21-channel multimodal input (Satellite, Digital Elevation Models, Water Indices)  for water segmentation under limited data. 
    \item Comparison between supervised baseline and self-supervised foundation model was conducted under limited data conditions.
    \item Supervised model demonstrated better effectiveness, while SSL improves cross-region performance stability.
    \item The proposed damage assessment workflow produced results that closely matched official post-event reports.

\end{itemize}

\newpage

\section{Introduction}\label{sec1}

Floods, being one of the most destructive natural disasters, cause significant damage to infrastructure, ecosystems and human life~\cite{brody2014examining, patil2020survey}. Each year, floods cause more than \$40 billion in damage to the world economy~\cite{douris2021atlas} and only for 2023 about 20 million people were affected~\cite{van2024global}. In Russia, for example, floods in Yakutia and Tulun in 2018-2019 caused more than \$400 million in damages~\cite{tananaev2021assessment,Shalikovsky2019The}. Moreover, over the past three decades, floods are occurring more frequently every year~\cite{li2022review, Patrick2020FeasibilityAO} and flood prediction itself has become an increasingly challenging task due to globally observed alterations in river regimes. 

Flood monitoring systems are based on diverse approaches from classical ground-based surveys, which rely on weather forecasts and stationary river gauges~\cite{Kamau2024, PANGALISHARMA201918} to various remote sensing approaches~\cite{sadiq2023remote, rs14030690}. Ground-based surveying can be effective for local early warning, but these methods suffer from slow performance and lack of spatial coverage. Remote sensing has become an essential tool for assessing and managing flooded areas, providing critical data for water resources management and protection. This is particularly crucial in flood emergency situations, where it is used for mapping, rapid assessment, and response activities~\cite{chowdhury2017use, pandey2022extreme}. The use of remote sensing data is especially relevant for floods monitoring, as vast territories with diverse climatic conditions pose unique monitoring challenges and for some regions it is hardly possible to perform any informative ground-based monitoring~\cite{Zelentsov2018THEUO}. 

Recent advances in computer vision and deep learning have further improved floods monitoring performance~\cite{lee2024deep}. Deep learning models trained on satellite imagery can outperform algorithmic approaches such as index-based techniques~\cite{Xu20072006} or thresholding, due to their robustness and scalability~\cite{bentivoglio2022deep, Shastry2023MappingFF}. Nowadays global actors such as Google~\cite{Nearing2024GlobalPO, googlefloodhub2024}, have launched AI-driven flood forecasting services, however, operational deployment remains limited. Either annotated datasets are scarce and models trained on local regions often fail to generalize across diverse hydrological regimes, and irregular satellite availability hinders near-real-time monitoring~\cite{popandopulo2023flood} or access to data sources or monitoring systems can be limited for Russian territory. These limitations highlight the urgent requirement for monitoring systems that can remain robust in the face of heterogeneous data and regional constraints.

The growing diversity of Earth observation data provides unique opportunity for precise, near-real-time flood monitoring~\cite{Amitrano2024FloodDW, popandopulo2023flood}. Diverse data modalities are becoming available offering various opportunities for flood monitoring thanks to the different physical senses of the recorded data: Synthetic Aperture Radar (SAR), multispectral optical sensors, and global elevation datasets. However, although images are abundant, annotated labels for flood events remain scarce, unevenly distributed, and costly to produce~\cite{illarionova2026esg}. This imbalance limits the effectiveness of traditional supervised approaches, which struggle to generalise across regions and sensor modalities due to domain shifts and insufficient training data. Self-supervised learning (SSL) offers a promising alternative by enabling models to learn generalisable representations from unlabelled data that can be fine-tuned for specific downstream tasks. Recent work in remote sensing has emphasised the potential of SSL to significantly reduce dependence on annotated datasets. For example, contrastive learning methods such as Momentum Contrast (MoCo) and SimCLR have been successfully adapted for satellite imagery to improve representation learning across different modalities and resolutions~\cite{Bourcier2024LearningRO, wang2023ssl4eo}. More recently, masked autoencoders (MAEs) have gained popularity due to their effectiveness in reconstructing missing or masked parts of an image based on the surrounding context~\cite{he2022masked} and such approaches are used for diverse satellite-based models construction~\cite{Nakayama2024SatSwinMAEEA, Reed2022ScaleMAEAS}. The next stage in the evolution of the MAE approach was the development of the Joint Embedding Predictive Architecture (JEPA)~\cite{assran2023self}, which masks and reconstructs image embeddings, enabling the unnecessary or unpredictable pixel-level details to be ignored. As satellite data for the same region of interest can differ due to weather, time of day or acquisition angle JEPA-based approaches are becoming popular as they enable the avoidance of such data deviations~\cite{Li2023PredictingGI, Astruc2024OmniSatSM}.

Another problem is related to the growing diversity of available satellite data. Flood monitoring particularly benefits from multimodal data, as different sources provide complementary perspectives. Synthetic Aperture Radar (SAR) data, such as that from Sentinel-1, is valuable due to its all-weather, day-and-night imaging capabilities, which are critical during cloud-covered flood events~\cite{Fouad2022}. Multispectral imagery provides optical data for land cover analysis and water body delineation in clear conditions. Digital elevation models (DEMs) and their derivatives provide topographic context, thereby improving the accuracy of flood extent predictions by accounting for terrain characteristics. However, combining these modalities remains challenging due to differences in resolution, noise, and format. SSL approaches such as FUS-MAE~\cite{ChanToHing2024FUSMAEAC} and SeaMo~\cite{li2024seamo} address these challenges by learning joint representations across modalities, which can be fine-tuned for water surface detection. While SSL approaches such as contrastive learning, MAE, and JEPA show promising results for satellite data usage, their application to multimodal flood detection remains underexplored. Most studies to date have focused on dual-modal inputs, such as those in~\cite{Wang2023DeCURDC, li2024seamo} or triple-modal inputs, such as those~\cite{Astruc2024OmniSatSM}, with limited integration of topographic information, such as DEM or different data resolutions. Furthermore, little work has directly linked water surface detection with downstream socioeconomic damage assessment. Integrating such assessments into AI-driven pipelines is essential for converting technical outputs into practical insights for emergency management agencies.

This study introduces a comprehensive framework for flood monitoring and damage assessment. The study aims to evaluate and compare the effectiveness of supervised and self-supervised learning strategies for water surface segmentation in situations where data is limited. To this end, multimodal datasets combining radar, optical and topographic information were compiled for use in both pre-training and segmentation tasks. Based on the segmentation results, an approach for economic damage estimation in urban areas was developed in accordance with the official methodologies of the Russian Ministry of Emergency Situations. The proposed framework serves as a technical benchmark and practical tool, demonstrating how multimodal, data-efficient learning methods can improve the scalability and operational readiness of AI-driven flood monitoring in the Russian Federation.

\paragraph{Main contribution} 

\begin{figure*}[h]
    \centering
    \includegraphics[scale=0.35]{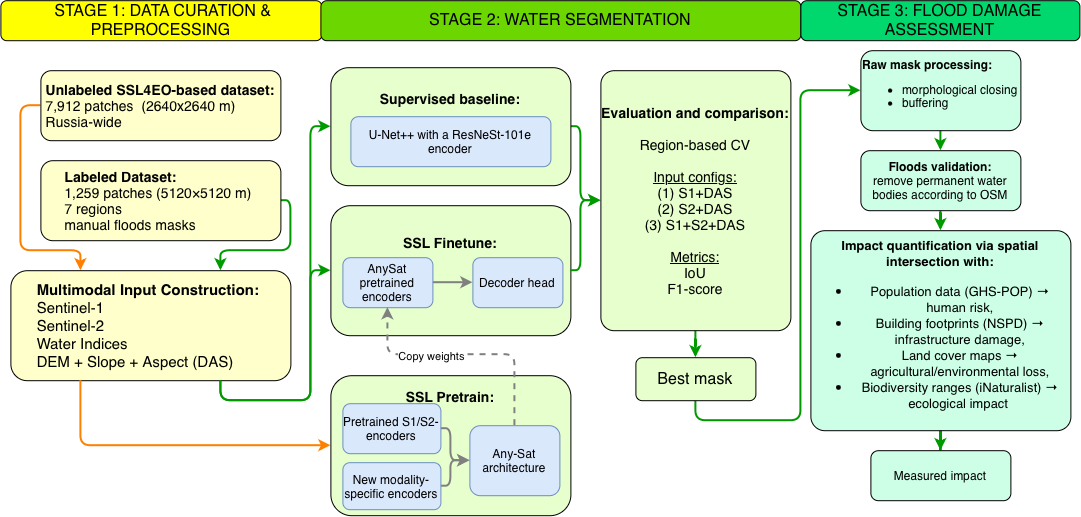}
    \caption{Framework pipeline}
    \label{fig:framework_pipeline_draft}
\end{figure*}

Current work main contribution is a structured comprehensive multi-stage pipeline construction. First, we develop a robust flood segmentation model, which we then apply to a practical damage assessment framework. The core of our approach is a comparative analysis between of a state-of-the-art self-supervised learning (SSL) model and a traditional supervised baseline. This methodology enables us to rigorously test the hypothesis that a foundation model pre-trained on a large, unlabelled dataset could outperform a standard model that has been trained solely on limited labelled data. The entire workflow is depicted in Figure~\ref{fig:framework_pipeline_draft} and detailed in the subsequent sections.

\section{Data Curation and Preprocessing} 
 
\subsection{Study Area}
To evaluate cross-regional generalisation of the proposed framework, we selected seven regions of the Russian Federation that jointly span the country's principal climatic, hydrological and flood-generation regimes (Table~\ref{tab:regions}, Figure~\ref{fig:region_data_distribution}). The selection covers a longitudinal transect of roughly 100$^{\circ}$ — from the East European Plain (Moscow region) and the semi-arid steppes of the Southern Urals (Orenburg region, including Orenburg and Orsk) in the west, through the West Siberian Plain (Kurgan region) and the sharply continental zone of Eastern Siberia (Krasnoyarsk Krai, Irkutsk region, Zabaykalsky Krai), to the maritime monsoon-influenced Far East (Sakhalin region). Together, these regions encompass the main flood-generation mechanisms operating in Russia: spring snowmelt and rain-on-snow events on plains and foothills, summer convective and frontal rainfall floods, and cyclone- and typhoon-driven extreme precipitation on the Far Eastern coast. All selected regions have been affected by significant recent flood events that motivated their inclusion — most notably the 2019 Iya River flood in Tulun~\cite{markin2020flooding}, the 2024 Ural River freshet and Orsk embankment failure in the Orenburg and Kurgan regions~\cite{orsk2024dam}, and recurring typhoon-related floods on Sakhalin such as the 1981 Phyllis and 2020 Maysak events~\cite{razjigaeva2020recurrence, medvedev2022destructive}. The regions also span a wide range of exposure conditions, from densely populated and infrastructure-rich landscapes (Moscow region) to sparsely populated areas with limited ground-based monitoring (Ushmun in Zabaykalsky Krai), which is essential for evaluating the downstream damage-assessment module.

\subsection{Dataset Composition}
The foundation of our work rests on two distinct datasets: (1) a large-scale unlabelled corpus for self-supervised pre-training, and (2) a region-specific annotated dataset for supervised segmentation and fine-tuning. Each dataset is formed from a combination of Sentinel-1 (S1), Sentinel-2 (S2) data and digital elevation models (DEM) and their derivatives: slope and aspect from DEM and water-related indices from S2, forming a 21-channel multimodal input. S1 provides dual-polarization (VV, VH) radar data along with derived ratio features, while S2 contributes 10 spectral bands (B2, B3, B4, B5, B6, B7, B8, B8a, B11 and B12).

Each modality provides unique and complementary information. S1 captures structural data through radar signals, S2 provides multispectral optical imagery, and DEM and its derivatives offer topographic context, which is critical for flood detection tasks. 

\paragraph{Indices.}

In addition to the raw spectral bands, we derived a set of five commonly used water-related indices from S2 multispectral data. 
These indices are widely applied in remote sensing to highlight open water surfaces, soil moisture, and vegetation water content. 
Their inclusion provides a useful reference baseline for flood extent analysis and complements the deep learning models used in this study.

\begin{itemize}
    \item \textbf{NDWI (Normalised Difference Water Index)} emphasises the presence of water bodies and surface moisture by exploiting the difference between the green and near-infrared (NIR) bands~\cite{mcfeeters1996use}.
    \item \textbf{MNDWI (Modified Normalised Difference Water Index)} modifies the NDWI by replacing NIR with the shortwave infrared (SWIR) band. This reduces the effect of dense vegetation and makes MNDWI more reliable for detecting open water surfaces~\cite{xu2006modification}.
    \item \textbf{SWI (Standardised Water-Level Index)} estimates soil moisture conditions by contrasting reflectance in the NIR and SWIR ranges~\cite{Bhuiyan2004VariousDI}.
    \item \textbf{AWEI$_{sh}$ (Automated Water Extraction Index, shadow version)} uses a weighted combination of green, NIR and SWIR bands to improve water delineation accuracy, particularly in shadowed or complex environments~\cite{Feyisa2014AutomatedWE}.
    \item \textbf{AWEI$_{nsh}$ (Automated Water Extraction Index, non-shadow version)} provides an alternative formulation designed for cases where SWIR coverage is limited, offering a more robust detection under such conditions~\cite{Feyisa2014AutomatedWE}.
\end{itemize}

The indices are defined as follows:
\begin{equation}
NDWI = \frac{\rho_{green} - \rho_{NIR}}{\rho_{green} + \rho_{NIR}}
\end{equation}

\begin{equation}
MNDWI = \frac{\rho_{green} - \rho_{SWIR}}{\rho_{green} + \rho_{SWIR}}
\end{equation}

\begin{equation}
SWI = \frac{\rho_{NIR} - \rho_{SWIR}}{\rho_{NIR} + \rho_{SWIR}}
\end{equation}

\begin{equation}
AWEI_{sh} = 4 \times \rho_{green} - (0.25 \times \rho_{NIR} + 6.75 \times \rho_{SWIR})
\end{equation}

\begin{equation}
AWEI_{nsh} = \rho_{green} + \rho_{red} - 2 \times \rho_{SWIR}
\end{equation}

\paragraph{Self-Supervised pre-training data.}
For the self-supervised learning (SSL) stage, we curated a multimodal, multitemporal dataset from the publicly available SSL4EO-S12 dataset~\cite{wang2023ssl4eo}. S1, S2 images were selected within the latitude range of \textit{45°N–82°N} and the longitude range of {30°E–180°E} to cover the Russian territory. Totally 7,912 unique geographic locations were selected 2,640$\times$2,640\,m each. We extracted the geographic coordinates for each location and downloaded the corresponding digital elevation model (DEM) data at a 30-meter resolution from open sources, calculated aspect, slope and water-related indices. The SSL4EO dataset also provides four images per modality taken at different times to increase the stability of the model's performance in relation to seasonal and daytime deviations. This produced a 21-channel, multimodal, multitemporal input with a spatial resolution of 10–30 m.

\paragraph{Sentinel metadata description}
The file names of S1 and S2 contain metadata that provides information about Earth observations received from these satellites. For S1, the standard file naming scheme includes data such as: 
\begin{itemize}   
    \item YYYYMMDDTHHMMSS - represents the start and end timestamps of the survey in the format year-month-day-hour-minute-second (UTC).
    \item OOOOO is the orbit identifier (a five-digit code).
\end{itemize}

For S2, the naming scheme includes, in addition to the date, the TXXXXX tile code in the MGRS (Military Grid Reference System) system. This metadata enables the time and location of the survey to be determined accurately, making it possible to analyse and process satellite images.

\begin{figure*}[h]
    \centering
    \includegraphics[scale=0.4]{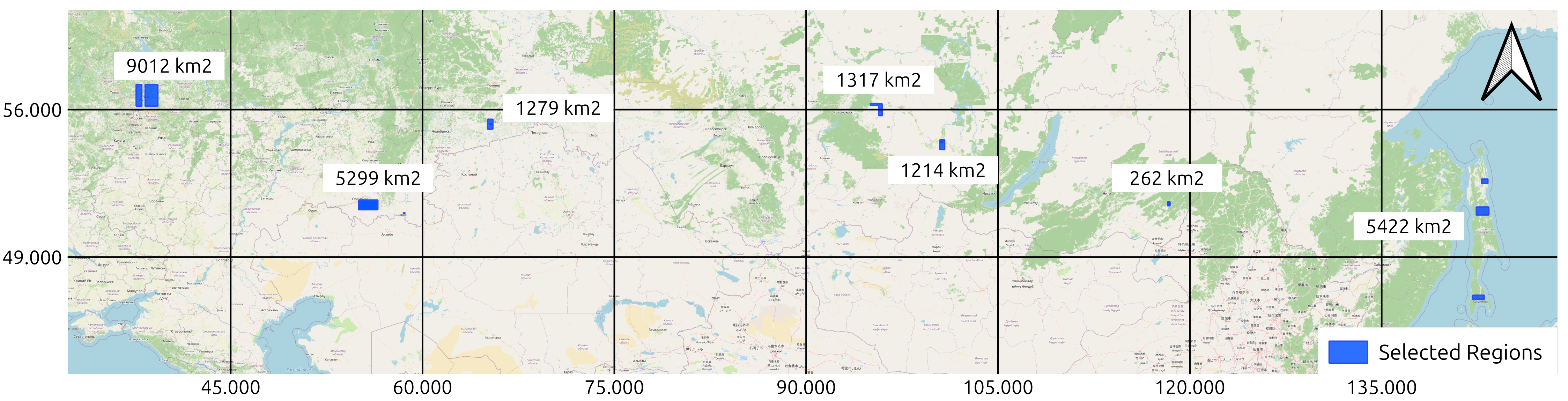}
    \caption{Selected regions data distribution and the corresponding encompassed area. The map was generated with the QGIS 3.34 software (https://qgis.org) and base map is configured using OpenStreetMap database (www.openstreetmap.org).}
    \label{fig:region_data_distribution}
\end{figure*}

\begin{figure*}[!htbp]
    \centering

    \includegraphics[scale=0.4]{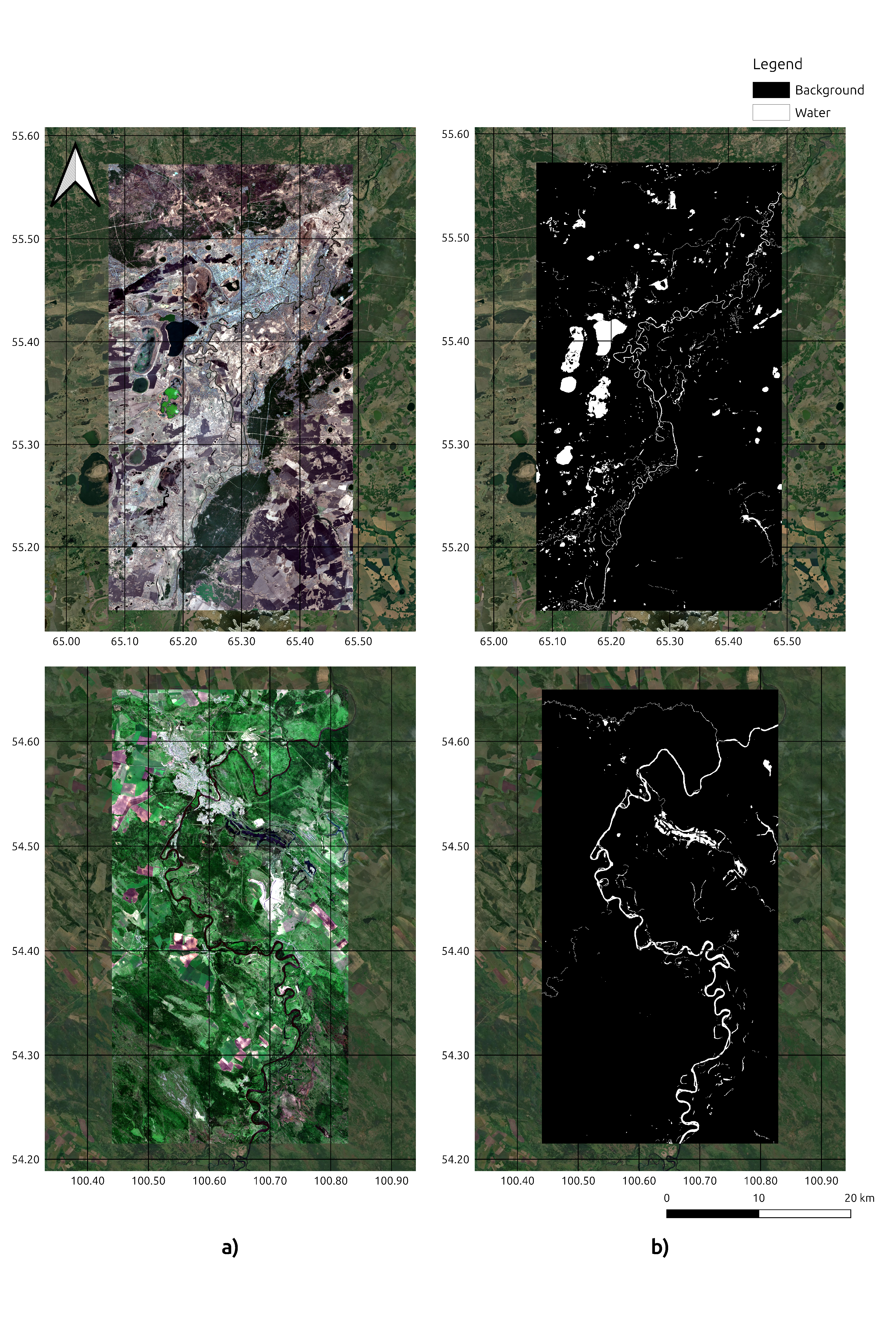}
    \caption{Examples of flood annotation for self-collected data: \textbf{a)} RGB channels from the Sentinel-2 satellite, \textbf{b)} manual original markup of the Sentinel-2 data. The map was generated with the QGIS 3.34 software (https://qgis.org) and base map from the database Mapbox (https://www.mapbox.com)}
    \label{fig:label_examples}
\end{figure*}

\paragraph{Supervised segmentation data.}
For fine-tuning and evaluation, self-collected and annotated dataset is used, covering seven regions of Russia and encompassing diverse hydrological and geographic conditions~(Data distribution and characteristics are presented in table~\ref{tab:regions} and figure~\ref{fig:region_data_distribution}). In each region, flood masks were manually labelled at pixel level to distinguish between flooded and non-flooded areas. Region visualisation and the following labels are presented in Figure~\ref{fig:label_examples} for the Kurgan and Tulun regions. Initially downloaded images were split into 1259 image patches of 512$\times$512 pixels, ($\sim$5$\times$5 km) with 64-pixel overlap. Each patch includes aligned S1, S2, and DEM inputs. DEM and S2 derivatives (slope and aspect, water-related indices) were added, forming the DAS feature set (DEM, aspect, slope) with additional index information. This dataset provides the ground truth for benchmarking supervised and SSL-based approaches. The S1 and S2 data has 10 m spatial resolution, and the DAS data were initially obtained in 30 m resolution and up-scaled to 10 m using bicubic interpolation. 

\begin{table}[h]
\centering
\caption{Overview of selected regions: Each region contributes with manually annotated flood mask, aligned S1 and S2 data, and DAS index inputs.}
\begin{tabular}{|l|l|l|}
\hline
Region                           & Area, km$^{2}$ & Patches \\ \hline
Krasnoyarsk krai (Kansk)       &   1317   & 83     \\
Kurgan region (Kurgan)           &   1279   & 66     \\
Moscow region                    &   9012   & 460     \\
Orenburg region (Orenburg, Orsk) &   5299   & 266     \\
Irkutsk region (Tulun)           &   1214   & 66      \\
Sakhalin region                  &   5422   & 288     \\
Zabaykalsky krai (Ushmun)        &   262   & 30      \\ \hline
\end{tabular}
\label{tab:regions}
\end{table}

\section{Methods. Water Segmentation and Comparative Analysis} 
The framework's main focus is on water surface detection for flooded territories monitoring and damage assessment. The problem of data scarcity led to the following solution: several separate water surface detection models were developed to deal with data scarcity problem. Depending on the available images (S1, S2, S1 + S2) one of these models will be selected. As the models must be robust enough to operate in different Russian regions, two approaches were considered: supervised and self-supervised. The supervised approach is considered as a baseline and we utilised a U-Net++ architecture~\cite{zhou2018unet++}. For the SSL approach AnySat~\cite{astruc2024anysat} was considered as a promising tool for working with multimodal data. The core of our investigation is a direct comparison between the SSL-based approach and the traditional supervised baseline. 

\subsection{Self-Supervised Learning Approach} 

Firstly, we provide an overview of the AnySat architecture~\cite{astruc2024anysat}. AnySat is a versatile Earth observation model designed to handle multimodal data across diverse resolutions, scales, and modalities. The architecture is based on a self-supervised learning framework called Joint Embedding Predictive Architecture (JEPA)~\cite{assran2023self}, which is a representative of feature predictive SSL architectures. 
The objective is to learn how to reconstruct the masked sections of input images within the feature space. To achieve this, JEPA employs a context encoder to convert masked images into an abstract representation, followed by a predictor module that predicts the representation generated by a target encoder receiving an unmasked input. Compared to the input (pixel) space, the feature space allows to ignore unnecessary or unpredictable pixel-level details in the target representation such as weather, time of day, or acquisition angle. The AnySat approach uses MLP-based scale-adaptive spatial encoders for both the context and target encoders. This allows the model to operate with multimodal inputs that have diverse data resolution, scale and modalities. This design enables AnySat to support heterogeneous sensors simultaneously, including optical imagery (e.g. aerial and SPOT), multispectral time series (e.g. S2 and Landsat) and radar data (e.g. S1 and ALOS), without the need for image resizing. This provides a reusable backbone for downstream tasks.

\paragraph{Data input processing}

We follow the AnySat tiling strategy so that all modalities can be aligned on a shared patch grid. Let $S$ denote a square tile of side length $S$ meters and let $P$ be the patch size (in meters) shared across modalities.  For modality $m$, let $R_m$ the spatial resolution (meters per pixel),$T_m$ the temporal depth ($T_m{=}1$ for single-date), and $C_m$ the channel count. A patch therefore spans
\[
\Delta_m \;=\; P / R_m.
\]
Each patch $x^m_p \in \mathbb{R}^{\Delta_m \times \Delta_m \times T_m \times C_m}$ is subdivided into fixed-size sub-patches of $\delta_m \times \delta_m$ pixels, yielding $(\Delta_m/\delta_m)^2$ sub-patch tokens per patch. 
A \emph{modality-specific projector} $\phi^{\mathrm{proj}}_m$ maps each flattened sub-patch to an $E$-dimensional vector,
\[
\phi^{\mathrm{proj}}_m:\; \mathbb{R}^{(\Delta_m/\delta_m)^2\delta_m^2\,T_m\,C_m} \rightarrow \mathbb{R}^{(\Delta_m/\delta_m)^2E},
\]
after which a \emph{shared spatial transformer} $\phi^{\mathrm{trans}}$ aggregates the sequence of sub-patch embeddings into a single patch representation $f^{\,m}_p$. Since $\delta_m$ is fixed per modality, changing $P$ only varies the number of input tokens to $\phi^{\mathrm{trans}}$—the output dimensionality $E$ remains constant across modalities and patch sizes. Finally, a \emph{modality combiner} $\phi^{\mathrm{comb}}$ fuses $f^{\,m}_p$ from all modalities into one multimodal token $f^{\star}_p$ per patch via cross-attention, providing a metrically consistent interface for heterogeneous sensors without any image resizing.

\paragraph{Pre-training stage}

\begin{figure*}[!htbp]
    \centering
    \includegraphics[scale=0.1]{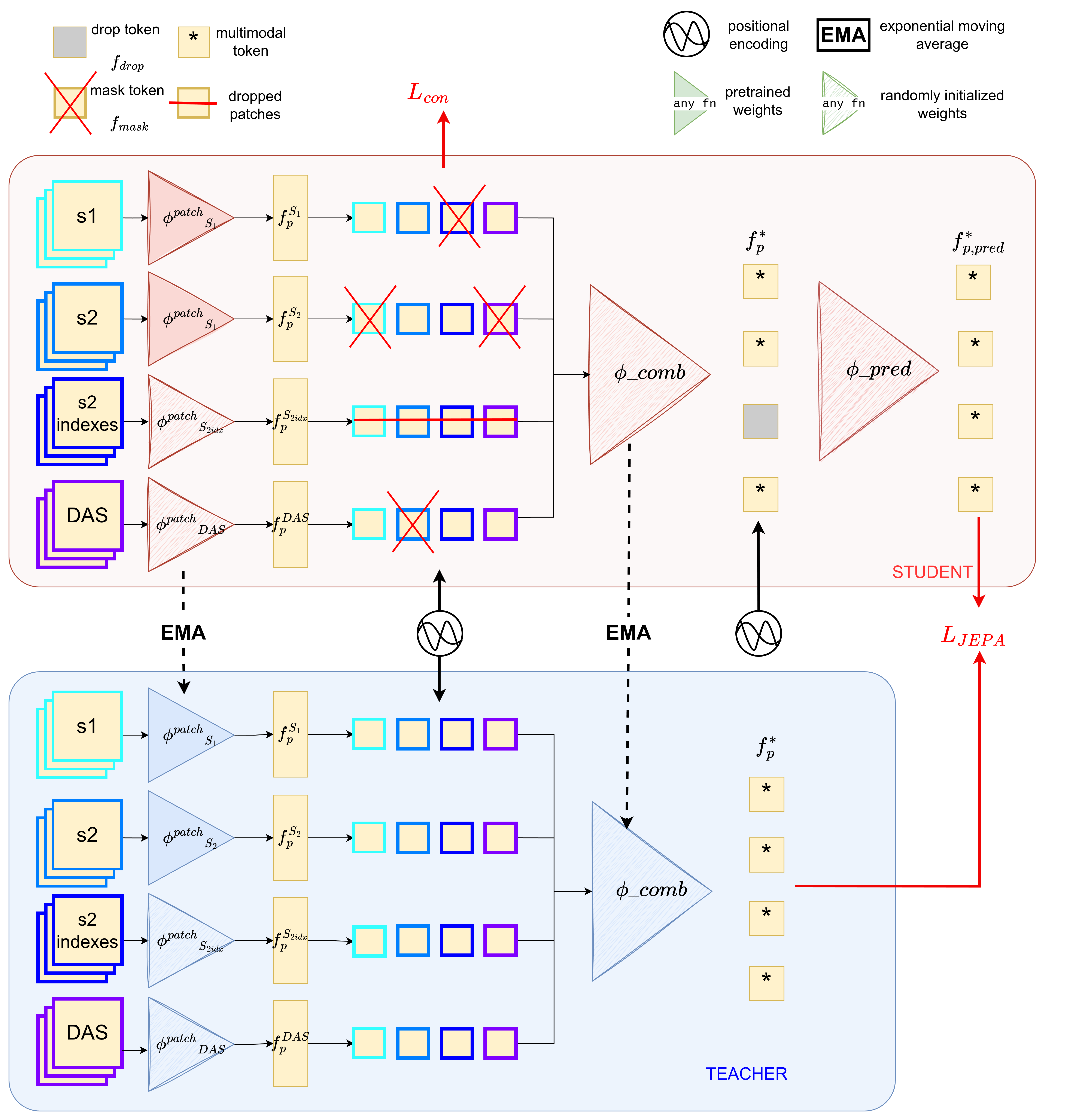}
    \caption[AnySat architecture]{%
  \textbf{AnySat-based architecture used in this work.}
  The diagram shows a \emph{student} (top) and a \emph{teacher} (bottom) network.
  For each modality $m$ (S1, S2, S2-originated indices, and DAS: DEM with aspect/slope), a \emph{patch encoder} $\phi^{\text{patch}}_{m}$ maps visible patches to per-patch tokens $f^{m}_{p,\mathrm{S/T}}$.
  Then dropping and masking are applied only to the student inputs: some patches have all modalities removed (dropping), while others have only random modalities removed (masking).
  The remaining tokens are fused by the \emph{modality combiner} $\phi^{\text{comb}}$ into a \emph{multimodal} token $f^{\star}_{p,\mathrm{S/T}}$. 
  The student additionally contains a \emph{predictor} $\phi^{\text{pred}}$ that produces $f^{\star}_{p,\mathrm{pred}}$ for masked patches.
  Training minimizes (i) a \textbf{JEPA loss} $\mathcal{L}_{\mathrm{JEPA}}$ that aligns $f^{\star}_{p,\mathrm{pred}}$ with $f^{\star}_{p,\mathrm{T}}$ at masked locations, and (ii) a \textbf{Cross-modal contrastive loss} $\mathcal{L}_{\mathrm{con}}$ that aligns co-located per-modality tokens across visible modalities. 
  The teacher parameters are an \emph{exponential moving average} (EMA) of the student. 
  Initialization follows our reuse strategy: the S1/S2 encoders are loaded from public AnySat checkpoints (``pre-trained weights''), whereas the S2-indices and DAS branches start from random initialization.
  }
    \label{fig:anysat architecture}
\end{figure*}

Since AnySat architecture is based on the JEPA paradigm, it involves two networks that are used during the training: a student and a teacher (Figure~\ref{fig:anysat architecture}). For both networks, co-registered inputs are partitioned into spatially aligned patches and processed as described above (projectors $\phi^{\mathrm{proj}}_m$ $\rightarrow$ shared transformer $\phi^{\mathrm{trans}}$ $\rightarrow$ combiner $\phi^{\mathrm{comb}}$). During SSL, the student network processes inputs with significant masking and uses a lightweight predictor head to reconstruct the teacher`s latent patch embeddings for the hidden patches. Patch-level masking and dropping are applied uniformly across modalities; differences in native pixel resolution are absorbed by the scale-adaptive encoder, avoiding any image resizing.
The teacher network is updated as an Exponential Moving Average (EMA) of the student's weights~\cite{he2020momentum}.

The model is trained using a combination of two losses:
\begin{itemize}
\item \textbf{Cross-modal contrastive loss:} an InfoNCE objective that aligns co-located patch embeddings across different modalities. 
\item \textbf{JEPA loss:} an $L_2$ alignment between student`s predictions and teacher`s embeddings in the masked regions;
\end{itemize}

\paragraph{Fine-tuning stage.} 

\begin{figure*}[h]
    \centering
    \includegraphics[scale=0.25]{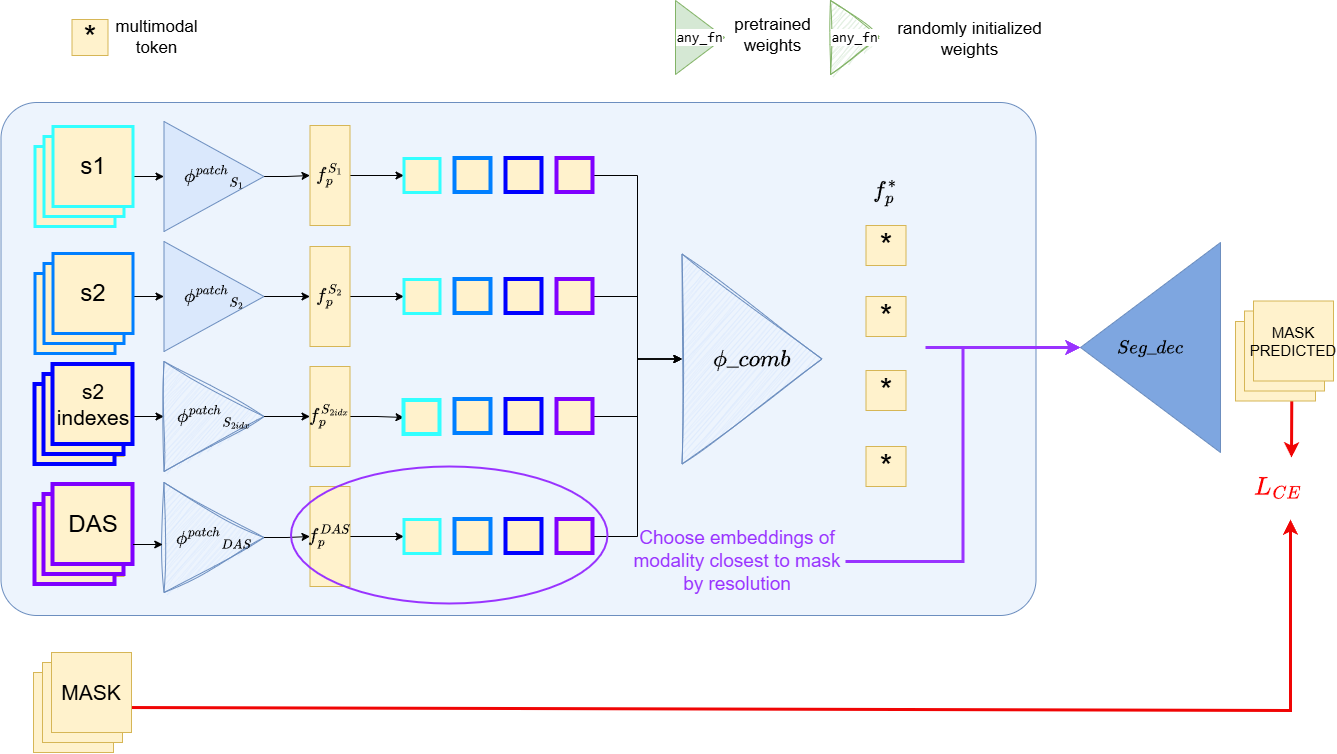}
    \caption[Downstream task pipeline]{%
    \textbf{Downstream task pipeline.}
    We reuse the SSL-adapted backbone (modality-specific patch encoders $\phi^{\mathrm{patch}}_{m}$ and combiner $\phi_{\mathrm{comb}}$) to produce per-modality tokens $f^{m}_{p}$ and a multimodal token $f^{\star}_{p}$.
    For dense prediction, we choose a \emph{reference modality} whose native spatial resolution best matches the annotation grid. 
    Its sub-patch embeddings are concatenated with the corresponding multimodal token $f^{\star}_{p}$ and fed to a lightweight decoder to obtain a pixel-wise flood mask on the reference grid.
    Training minimizes a weighted cross-entropy loss $\mathcal{L}_{\mathrm{CE}}$ between the predicted and ground-truth masks.
    }
    \label{fig:anysat downstream}
\end{figure*}

We reuse the SSL-adapted backbone from the teacher network (modality projectors, shared transformer, and combiner) and attach a lightweight segmentation head to produce dense predictions at the target label resolution (see Figure ~\ref{fig:anysat downstream}). Following AnySat, we select a reference modality whose native resolution is closest to the annotation grid. For each patch, we form a high-resolution dense feature map at the sub-patch scale by concatenating (i) the reference modality's sub-patch embeddings (outputs of the modality-specific projector) with (ii) the corresponding multimodal patch token (output of the combiner). The segmentation head then projects these merged sub-scale and patch-scale representations to per-pixel logits on the sub-patch grid. This sub-patch conditioning yields finer predictions than relying on patch-level tokens alone. Training uses a weighted cross-entropy loss with class balancing.

\subsection{Experimental Protocol and Metrics} 

To ensure a fair and robust comparison, both pathways are evaluated under an identical experimental setup. Each model was trained and tested on three different input configurations: (1) S1 + DAS, (2) S2 + DAS, and (3) S1 + S2 + DAS. The evaluation is performed using a strict region-based cross-validation scheme, where data from entire geographic regions are held out for testing. This prevents spatial data leakage and provides a more realistic measure of the models' ability to generalise to unseen areas. 

Model performance is assessed using standard semantic segmentation metrics, namely \textit{Intersection over Union} (IoU, also referred to as the Jaccard index) and the \textit{F1-score} (defined as the harmonic mean of precision and recall). These metrics are widely used to quantify segmentation accuracy and are computed as follows:

\begin{equation}
\mathrm{Precision} = \frac{TP}{TP + FP},
\end{equation}
\begin{equation}
\mathrm{Recall} = \frac{TP}{TP + FN},
\end{equation}
\begin{equation}
F1\text{-score} = 2 \times \frac{\mathrm{Precision} \times \mathrm{Recall}}{\mathrm{Precision} + \mathrm{Recall}},
\end{equation}
\begin{equation}
IoU = \frac{TP}{TP + FP + FN},
\end{equation}

\noindent where $TP$ (true positives) is the number of pixels correctly classified, $FP$ (false positives) is the number of pixels incorrectly predicted as belonging to the target class, and $FN$ (false negatives) is the number of target-class pixels that were incorrectly assigned to another class. To provide a comprehensive evaluation, both IoU and F1-score were computed in ``macro'' mode, i.e., ground-truth and predicted labels were concatenated across each test image into single vectors before metric calculation. This ensures that the reported values reflect average model performance over individual images. Another metric that could be considered is the Accuracy score, which measures the ratio of correctly classified pixels to the total number of pixels. However, the current metric is unable to demonstrate informative results under the class imbalance conditions; that's why it is not preferable in this study.

\section{Flood Damage Assessment} 

In the final stage, the water segmentation masks generated by the best-performing model from Stage 2 are used as the primary input for damage estimation. The design of this framework is explicitly aligned with the regulatory guidelines established by the official Russian authority for emergency response, Ministry of Emergency Situations of Russia (EMERCOM).
The methodological foundation for damage assessment is prescribed by the EMERCOM order ``Methodology for Assessing Damage from Emergency Situations''~\cite{emercom2020}. In accordance with this directive, the aggregate damage is quantified as an absolute measure of the total harm inflicted upon the following components:
\begin{enumerate} \item \textbf{Risk to Life and Health, and private property:} assessment of the amount of damage to human life and health, property of individuals in terms of essential property, as well as immovable property; \item \textbf{Damage to Government and Municipal Property:} evaluation of losses incurred by state and municipal institutions; \item \textbf{Ecological and Environmental Damage:} Comprehensive assessment of harm to the environment, including: damage to animal and plant life; impact on surface and groundwater; degradation of forests and natural objects; pollution of atmospheric air and soil; and adverse effects on flora and fauna species. \end{enumerate}

\paragraph{Damage assessment methodology:} Numerous approaches exist for utilizing remote sensing data in flood damage assessment~\cite{kucharczyk2021}. Many studies focus on land cover classification through machine learning models~\cite{rahman2020}\cite{billah2023} or by relying on satellite-derived indices~\cite{sajjad2020}. However, these methods typically lack the granularity to identify specific infrastructure elements, such as individual buildings or roads, and remain focused on general land cover categories~\cite{billah2023}. Consequently, they often serve primarily to pre-process information for subsequent human-led damage evaluation.

For the territory of the Russian Federation, which offers extensive open data and cadastral information from state systems, it is possible to circumvent these limitations. The integration of these authoritative datasets enables more efficient and data-rich environmental analysis, enhancing the accuracy of damage assessments.

The proposed framework operates by sequentially intersecting a water mask with various supplementary datasets to assess the impact on human populations, property, and the environment.
\paragraph {Raw mask processing:} The initial water segmentation mask requires morphological processing to account for inherent limitations in satellite-based flood detection. Infrastructure may be partially submerged yet remain visible from orbit if protruding above water level. To eliminate this problem the mask undergoes binary closing operation to connect disjointed bodies and mitigate under-segmentation. Subsequent buffering expands the flood boundary to compensate for two key factors: potential temporal misalignment between image acquisition and peak flood phase, and the extended indirect impact zone of the floodwaters (e.g., saturated soils, inaccessible infrastructure, and disrupted transport routes). 
\paragraph{Flood detection validation:} To distinguish actual flood events from permanent water bodies, the processed mask is cross-referenced with baseline hydrography data from OpenStreetMap (OSM)~\cite{openstreetmap}. A flood is confirmed if the exceeded water area (i.e., mask area minus permanent water bodies) surpasses a predetermined threshold. This stage also calculates the total inundated area for subsequent damage quantification.
\paragraph {Risk for Human Life and Health:} The risk to human life was estimated using the Global Human Settlement Population dataset (GHS-POP R2023A version 1.0, resolution from 3 to 30 arcseconds)~\cite{carioli2023}. This dataset provides global population estimates at 5-year intervals from 1975–2020, with projections for 2025 and 2030. Population counts within the inundated area were calculated by spatially intersecting the population raster with the floodwater mask and summing the values of affected pixels.

\paragraph {Infrastructure analysis:} The framework quantifies impacts on the built environment by intersecting the flood mask with building footprints from Russia’s National Spatial Data System (NSPD)~\cite{nspd2025}. The latter provides critical attributes including: building purpose, ownership status, official designation, and cultural heritage protection status. These attributes enable stratified damage accounting across residential, private, and state-owned buildings, with special categorization for socially critical infrastructure (schools, hospitals, kindergartens) and electrical substations.
\paragraph {Land cover impact quantification:} Utilizing a pre-developed land cover classification algorithm~\cite{Makarov2025}, inundated areas are categorized into: open water, settlements, bare soil, forest, grassland, agricultural land, and swamps. The total flooded area is disaggregated by these classes to assess environmental and agricultural damage, providing a comprehensive impact overview across natural and human-modified landscapes. 
\paragraph {Ecological impact:} The ecological impact of flooding on fauna and flora was assessed using the iNaturalist Open Range Map Dataset~\cite{inaturalist2025}, which provides georeferenced information on the global distribution of living organisms. The methodology involved a spatial intersection the water mask and the species range maps for key taxonomic groups: Actinopterygii, Amphibia, Arachnida, Chromista, Fungi, Mammalia, Animalia, Plantae, and Reptilia. For each group, we quantified the impact by counting the number of taxa whose habitats overlapped with the flooded area. Additionally, a list of the specific taxon names affected by the inundation was compiled. 

\section{Experimental Setup}

\subsection{Supervised Learning Approach (Baseline)}

\subsubsection{Data Preprocessing}

For each of the three data configurations (S1+DAS, S2+DAS, S1+S2+DAS), we trained distinct combinations of input channels. In order to ensure comparability across inputs, we applied a per-channel min-max normalization, using global minimum and maximum values that were computed in advance from the entire dataset for each channel:

\begin{equation}
\mathrm{MINMAX}(s[i]) = \frac{s[i] - globalmin[i]}{globalmax[i] - globalmin[i]},
\end{equation}

\noindent where $s[i]$ denotes the $i$-th channel of $s$, and \textit{globalmin} and \textit{globalmax} represent arrays containing the global per-channel minimum and maximum values, respectively. Aspect values were converted from degrees to radians using \textit{cosine} function.

\subsubsection{Model Architecture}

We implemented a supervised segmentation baseline using the U-Net++~\cite{zhou2018unet++} architecture coupled with a \texttt{ResNeSt-101e}~\cite{zhang2020resnest} encoder pre-trained on ImageNet~\cite{deng2009imagenet}. This backbone, with approximately 85.6 million trainable parameters, captures hierarchical representations across five depth levels, while the decoder reconstructs fine-resolution segmentation maps with progressively refined channel dimensions \([512, 256, 128, 64, 32]\).

\subsubsection{Training Setup}

Training employed the AdamW optimizer (learning rate $5 \times 10^{-4}$, weight decay $1 \times 10^{-4}$), in combination with a \linebreak \texttt{ReduceLROnPlateau} scheduler monitoring validation Intersection-over-Union (IoU). To address class imbalance, we used Focal Loss with parameters $\gamma = 2.0$ and $\alpha = 1/15$. Models were trained for up to 100 epochs with a batch size of 4 and early‑stopping on validation IoU. Data augmentation was applied to enhance robustness, including random 90° rotations, horizontal and vertical flips, brightness and contrast adjustments for S2 channels, and mild Gaussian and multiplicative noise.  

We perform 4‑fold cross‑validation with group‑wise splits to avoid spatial/scene leakage (folds defined at the geographic region level). We report fold‑wise validation means and cross‑fold mean $\pm$ standard deviation; full per‑fold tables with sample counts are provided in Table~\ref{tab:cv_splits}. This approach was chosen to assess better the robustness of water surface detection across distinct landscapes and acquisition conditions. For these splits we report results only on images containing more than 5\% water coverage to avoid trivial cases with negligible water extent.

\begin{table*}[!htbp]
\centering
\caption{Cross-validation splits showing the geographic regions assigned to training and validation, along with the number of images in each set.}
\footnotesize
\begin{tabular}
{|c|p{0.28\textwidth}|p{0.14\textwidth}|p{0.18\textwidth}|p{0.18\textwidth}|}

\hline
\textbf{Split} & \textbf{Training Regions} & \textbf{Validation Regions} & \textbf{Training Images} (with Area in km$^{2}$) & \textbf{Validation Images} (with Area in km$^{2}$) \\
\hline
1 & Kansk, Kurgan, Moscow, Orenburg, Orsk, Tulun & Sakhalin, Ushmun & 941 (18121 km$^{2}$) & 318 (5684 km$^{2}$) \\
2 & Kansk, Moscow, Sakhalin, Tulun, Ushmun & Kurgan, Orenburg, Orsk & 927 (17227 km$^{2}$) & 332 (6578 km$^{2}$) \\
3 & Kurgan, Moscow, Orenburg, Orsk, Sakhalin, Ushmun & Kansk, Tulun & 1110 (21274 km$^{2}$) & 149 (2531 km$^{2}$) \\
4 & Kansk, Kurgan, Orenburg, Orsk, Sakhalin, Tulun, Ushmun & Moscow & 799 (14793 km$^{2}$) & 460 (9012 km$^{2}$) \\
\hline
\end{tabular}
\label{tab:cv_splits}
\end{table*}

\subsection{Self-Supervised Learning Approach}

\subsubsection{AnySat architecture modifications}
\label{subsec:AnySat architecture modifications}

Originally, the AnySat architecture was not designed to handle SSL4EO-S12 dataset, which includes DAS and five additional S2-based water-related indices. We addressed this by developing new encoders for the DAS modality and incorporating the five S2-originated indices into the architecture. These new components maintain the same interface to enable seamless fusion in the combiner. 

\subsubsection{Data preprocessing}

Table~\ref{tab:modalities} summarises the sensing characteristics used during SSL pre-training and downstream fine-tuning. The sample size $S$ corresponds to the spatial support of each tile, $P$ denotes the base patch size retained in the pipeline, and the $(R, T, C)$ triple lists the spatial resolution, number of temporal observations, and channel dimensionality for each modality.

\begin{table}[h!]
\centering
\footnotesize
\setlength{\tabcolsep}{2pt}
\caption{Modalities, spatial supports, and resolutions used during SSL pre-training and fine-tuning.}
\begin{tabular}{lcccccc}
\toprule
\textbf{Modality} & \textbf{Sample (S)} & \textbf{Patch (P)} & \textbf{Stage} &
\multicolumn{3}{c}{\textbf{Resolution}} \\
\cmidrule(lr){5-7}
& & & & \textbf{Spatial (R)} & \textbf{Temporal (T)} & \textbf{Spectral (C)} \\
\midrule
DAS & \multirow{4}{*}{2640\,m} & \multirow{4}{*}{100\,m} & \multirow{4}{*}{SSL pre-training} & 20\,m & 1 & 3 \\
S1  &                         &                        &                                   & 10\,m & 4 & 3 \\
S2  &                         &                        &                                   & 10\,m & 4 & 10 \\
Water indices &               &                        &                                   & 10\,m & 4 & 5 \\
\midrule
DAS & \multirow{4}{*}{5120\,m} & \multirow{4}{*}{160\,m} & \multirow{4}{*}{Fine-tuning} & 10\,m & 1 & 3 \\
S1  &                         &                       &                              & 10\,m & 1 & 3 \\
S2  &                         &                       &                              & 10\,m & 1 & 10 \\
Water indices &               &                       &                              & 10\,m & 1 & 5 \\
\bottomrule
\end{tabular}
\label{tab:modalities}
\end{table}

We need to explain why the initial 30\,m DAS data is provided to the SSL model at a 20\,m resolution. Due to the model's architecture, it can only process square images. Since the DEM data was downloaded in the CRS EPSG:4326 projection, the pixel resolution of the images is non-uniform and depends on the longitude and latitude of the study area. To obtain square images, we interpolate the data, which changes the resolution from 30\,m to 20\,m.

The AnySat modality encoders are designed to work with data after standardization, which was also performed. The standardization for each channel \( x \) is calculated as:

\begin{equation}
x_{\text{standardized}} = \frac{x - \mu}{\sigma}
\end{equation}

where \( \mu \) is the mean and \( \sigma \) is the standard deviation of the channel. The mean and standard deviation were computed separately for each channel based on the training data.

For each stage we apply per-modality standardisation. During SSL pre-training we compute tile-wise means and standard deviations across the training split, aggregate them into global statistics, and use those to normalise every input. The fine-tuning stage repeats the same procedure on the fine-tuning training set so that downstream batches are re-centered with statistics matched to the supervised data distribution.

\subsubsection{Experiments}

We consider two evaluation settings to assess the model`s performance and compare results with baseline:
\begin{itemize}   
    \item \textbf{SSL pre-training $\rightarrow$ fine-tuning.} We reuse AnySat checkpoints for S1/S2 and then train in a self-supervised manner on our unlabelled dataset, then fine-tune on the annotated training data. 
    \item \textbf{Supervised training from scratch.} We randomly initialize all encoders and train directly on the labelled training set in a supervised manner.
\end{itemize}

\paragraph{Experiments 1: SSL pre-training and fine-tuning}

Since we use the same modalities (S1 and S2) as the original AnySat framework, we leveraged publicly available pre-trained AnySat weights for the S1 and S2 encoders. This preserves the existing feature extraction capabilities and only initialize the DAS and S2-originated indices encoders at random. This approach enables us to utilize pre-trained AnySat patch encoders for S1 and S2 data, avoiding the computational overhead of training them from scratch. Additionally, it allows new modalities to learn modality-specific sensors without disturbing the pre-trained optical/radar filters or requiring the large datasets typically needed for full model training.

\textit{SSL pre-training.} At this stage, we aimed to learn generalisable feature representations from our large, unlabelled SSL4EO-S12 dataset using AnySat’s self-supervised learning protocol. The complete model pre-trained for 12 epochs with a per-modality batch size 1. Actually, we curtailed the run at that point because each additional epoch would have exceeded the compute budget. The optimisation uses AdamW with a learning rate of $5 \times 10^{-5}$ and weight decay $1 \times 10^{-4}$ paired with a \linebreak \texttt{ReduceLROnPlateau} scheduler and an EMA decay coefficients in $(0.996, 1.0)$. We use 4 NVIDIA A100 for pre-training with roughly several GPU-hours per epoch and 60 GPU-hours in sum. 

\textit{Fine-tuning.} We attach a lightweight segmentation head and fine-tune it for supervised flood segmentation on the training split. For this, we use weighted cross-entropy with class balancing, AdamW with \linebreak \texttt{ReduceLROnPlateau}, early stopping based on validation loss (max 50 epochs). Fine-tuning takes approximately 2 hours to complete. 
To compare with the baseline model, we followed the same 4-fold cross-validation protocol. 

\paragraph{Experiment 2: Training from scratch}

To isolate the effect of SSL and pre-trained initialization, we trained an otherwise identical architecture from random initialization entirely. Training used the same configuration as in fine-tuning above: loss (weighted cross-entropy), optimizer (AdamW), scheduler (\texttt{ReduceLROnPlateau}), early stopping, etc.
For the comparison with the baseline, the same 4-fold cross-validation protocol was followed. 

\section{Results and Discussion}

\subsection{Water Surface Segmentation}

\subsubsection{Supervised Model Performance}

The trained Unet++ model, incorporating S1, S2, and DAS (elevation, aspect, slope) data, achieved promising segmentation results on these cross-validation sets (Tables~\ref{tab:aggregate_results_baseline} and \ref{tab:per_split_results_baseline}). Cross-validated experiments consistently demonstrated the advantage of multimodal fusion over single-source inputs (Table~\ref{tab:aggregate_results_baseline}). Averaged across four folds, the S1+S2+DAS configuration achieved the highest performance, with a mean IoU of $0.75 \pm 0.15$ and a mean F1-score of $0.84 \pm 0.11$. The S2+DAS model yielded slightly lower but comparably stable results (IoU $=0.74 \pm 0.15$; F1 $=0.83 \pm 0.11$), whereas S1+DAS was markedly less reliable, exhibiting a mean IoU of $0.57 \pm 0.29$ and strong fold-to-fold variability. 

Per-fold results further confirmed this ranking: for example, in Split~4, S1+S2+DAS achieved a mean IoU of 0.94 and F1-score of 0.97, while in Split~2, SAR-only models performed poorly (mean IoU $=0.15$, mean F1-score $=0.23$). Qualitative examples (Figures~\ref{fig:models_results_split1}, \ref{fig:models_results_split2}, \ref{fig:models_results_split3}, \ref{fig:models_results_split4}) further highlight that multimodal fusion enables robust delineation of inundated floodplains even in visually complex environments, whereas unimodal SAR models frequently misclassify noise and surface roughness.

\begin{table*}[!htbp]
\centering
\renewcommand{\arraystretch}{1.1}
\caption{Aggregate baseline performance across 4 cross-validation splits (images with $>$5\% water only).}
\begin{tabular}{@{}l c c c c@{}}
\toprule
\textbf{Modalities} & \textbf{Mean IoU} & \textbf{Mean F1} & \textbf{Median IoU} & \textbf{Median F1} \\
\midrule
   S1+DAS     & $0.57 \pm 0.29$ & $0.67 \pm 0.29$ & $0.57 \pm 0.34$ & $0.68 \pm 0.33$ \\
   S2+DAS     & $0.74 \pm 0.15$ & $0.83 \pm 0.11$ & $0.75 \pm 0.15$ & $0.85 \pm 0.09$ \\
   S1+S2+DAS  & $\mathbf{0.75 \pm 0.15}$ & $\mathbf{0.84 \pm 0.11}$ & $\mathbf{0.77 \pm 0.15}$ & $\mathbf{0.86 \pm 0.10}$ \\

\bottomrule
\end{tabular}
\label{tab:aggregate_results_baseline}
\end{table*}

\begin{table*}[!htbp]
\centering
\footnotesize
\renewcommand{\arraystretch}{1.1}
\caption{Per-split supervised model performance across training and validation sets, restricted to images with $>$5\% water.}
\begin{tabular}{@{}c l c c c c@{}}
\toprule
\textbf{Split} & \textbf{Modalities} & \textbf{Mean IoU} & \textbf{Mean F1} & \textbf{Median IoU} & \textbf{Median F1} \\
\midrule
1 & S1+DAS & 0.56 & 0.70 & 0.56 & 0.72 \\
  & S2+DAS & 0.65 & 0.77 & 0.63 & 0.77 \\
  & S1+S2+DAS & \textbf{0.66} & \textbf{0.78} & \textbf{0.65} & \textbf{0.78} \\
\addlinespace

2 & S1+DAS & 0.15 & 0.23 & 0.10 & 0.18 \\
  & S2+DAS & 0.59 & 0.72 & 0.63 & 0.77 \\
  & S1+S2+DAS & \textbf{0.60} & \textbf{0.73} & \textbf{0.64} & \textbf{0.78} \\

\addlinespace
3 & S1+DAS & 0.76 & 0.86 & 0.77 & 0.87 \\
  & S2+DAS & 0.78 & 0.88 & 0.81 & 0.89 \\
  & S1+S2+DAS & \textbf{0.80} & \textbf{0.89} & \textbf{0.83} & \textbf{0.91} \\

\addlinespace
4 & S1+DAS & 0.81 & 0.89 & 0.86 & 0.93 \\
  & S2+DAS & 0.93 & 0.96 & 0.93 & 0.96 \\
  & S1+S2+DAS & \textbf{0.94} & \textbf{0.97} & \textbf{0.94} & \textbf{0.97} \\

\bottomrule
\end{tabular}

\label{tab:per_split_results_baseline}
\end{table*}

\subsubsection{AnySat-based models performance}

Two types of experiments were conducted to evaluate the benefits of the self-supervised learning (SSL) approach. As it was described previously, multimodal segmentation models were initialized with SSL-pretrained encoders and subsequently fine-tuned for the water surface segmentation task. Their performance is summarized in Tables~\ref{tab:aggregate_results_ssl} and~\ref{tab:per_split_results_ssl}. For comparison, the same encoders were also trained from scratch to evaluate whether SSL pretraining provides measurable improvement (Tables~\ref{tab:aggregate_results_scratch} and~\ref{tab:per_split_results_scratch}).

Overall, SSL pretraining did not yield the expected performance gains. Across all modalities, SSL pre-training increases model performance stability, but decreases mean performance compared to from-scratch models. Although SSL aims to learn transferable spatial representations, in this case the pretrained features did not generalize effectively to the segmentation task. For Split~2 tangible improvements were observed. However, for the dataset as a whole, SSL pretraining did not provide tangible benefits and even caused dramatic degradation for the S1+DAS model. These findings indicate that supervised learning on labelled data is more effective than transferring SSL features for this task.

Among the approaches based on SSL pre-training S2+DAS modality combination demonstrated the best performance with mean IoU $=0.69 \pm 0.14$, mean F1-score $=0.79 \pm 0.12$. The S1+S2+DAS approach has comparably close results, however showing a little bit more performance stability. The S1+DAS results can be called unsatisfactory due to quite poor performance and big results deviation.

The same tendency is for from-scratch models: S2+DAS and S1+S2+DAS are showing relatively close to each other results with mean IoU $=0.71 \pm 0.15$, mean F1-score $=0.81 \pm 0.12$ and with mean IoU $=0.70 \pm 0.15$, mean F1-score $=0.81 \pm 0.12$ respectively. Figures~\ref{fig:models_results_split1},~\ref{fig:models_results_split2},~\ref{fig:models_results_split3},~\ref{fig:models_results_split4} demonstrate per-split AnySat models performance examples.

\subsubsection{Comparative Analysis}

When the AnySat-based pathway is compared with the supervised U-Net++ baseline, neither architecture dominates uniformly across all conditions. The AnySat-based SSL pretrained models matched or slightly outperformed U-Net++ on specific splits (on Split 2 across all modality combinations) suggesting that the scale-adaptive multimodal backbone can be beneficial when the test region exhibits significant distributional differences from the training regions. In aggregate, however, the U-Net++ baseline produced both higher mean accuracy and more consistent per-fold performance, and baseline water mask was therefore selected for the damage assessment pipeline.

The obtained results can be interpreted as a clear illustration that the optimal choice of architecture is conditional on data availability and task complexity. The U-Net++ baseline benefits from the availability of a powerful ImageNet-pretrained ResNeSt-101e encoder, which provides a strong inductive prior even with limited domain data. Conversely, the AnySat framework is designed to deliver its advantages in regimes that current setting does not fully reach. Specifically, several conditions appear to limit the realised benefits of the SSL pathway in this study:
\begin{itemize}
    \item \textbf{Pre-training corpus scale.} Foundation-style SSL is typically calibrated for unlabelled corpora orders of magnitude larger than the compiled dataset. At this scale, the DAS and index-specific encoders, which were initialised from scratch, are likely undertrained, and the pre-trained S1/S2 encoders cannot be substantially adapted to the local distribution.
    \item \textbf{Memory-constrained input geometry.} The AnySat implementation imposes GPU memory pressure that required reducing the effective input resolution during fine-tuning, which is known to penalise fine-grained, edge-sensitive segmentation tasks such as flood delineation.
    \item \textbf{Cross-modal feature alignment.} The SSL framework may have produced cross-modal representations that are not fully aligned between S1 and S2 in the current specific data distribution, limiting the gains of the S1+S2+DAS configuration relative to the spectrally dominant S2+DAS.
    \item \textbf{Domain gap between pre-training and fine-tuning data.} Differences in seasonality, surface composition and acquisition geometry between the SSL4EO-S12-derived pre-training tiles and the flood-event annotated tiles reduce the transferability of the learned embeddings.
\end{itemize}

For the operational scenario targeted by this study — water-surface segmentation over diverse Russian regions with a limited annotated dataset and a single downstream task — the supervised U-Net++ architecture is the more cost-effective and accurate choice, that will be propagated into the damage-assessment pipeline. The AnySat-based framework remains a promising direction when larger unlabelled multimodal corpora are available, robustness to systematically missing modalities at inference is operationally important (e.g. persistent cloud cover blocking S2), or a single backbone is to be reused across several downstream Earth observation tasks. Future work should therefore focus on scaling the pre-training corpus, mitigating the memory bottleneck, and evaluating AnySat in the missing-modality and multi-task regimes for which it was designed.

\begin{table*}[!htbp]
\centering
\renewcommand{\arraystretch}{1.1}
\caption{Aggregate SSL-based model performance across 4 cross-validation splits (images with $>$5\% water only).}
\begin{tabular}{@{}l c c c c@{}}
\toprule
\textbf{Modalities} & \textbf{Mean IoU} & \textbf{Mean F1} & \textbf{Median IoU} & \textbf{Median F1} \\
\midrule
  S1+DAS & $0.38 \pm 0.25$ & $0.49 \pm 0.25$ & $0.32 \pm 0.29$ & $0.43 \pm 0.29$ \\
  S2+DAS & $\mathbf{0.69 \pm 0.14}$ & $\mathbf{0.79 \pm 0.12}$ & $\mathbf{0.70 \pm 0.14}$ & $\mathbf{0.81 \pm 0.10}$ \\
  S1+S2+DAS & $0.68 \pm 0.11$ & $0.78 \pm 0.09$ & $0.69 \pm 0.09$ & $0.81 \pm 0.06$ \\
\bottomrule
\end{tabular}
\label{tab:aggregate_results_ssl}
\end{table*}

\begin{table*}[!htbp]
\centering
\footnotesize
\renewcommand{\arraystretch}{1.1}
\caption{Per-split SSL-approach performance across validation sets, restricted to images with $>$5\% water.}
\begin{tabular}{@{}c l c c c c@{}}
\toprule
\textbf{Split} & \textbf{Modalities} & \textbf{Mean IoU} & \textbf{Mean F1} & \textbf{Median IoU} & \textbf{Median F1} \\
\midrule
1 & S1+DAS & 0.28 & 0.42 & 0.23 & 0.37 \\
  & S2+DAS & 0.51 & 0.61 & 0.48 & 0.64 \\
  & S1+S2+DAS & \textbf{0.54} & \textbf{0.67} & \textbf{0.60} & \textbf{0.75} \\
\addlinespace
2 & S1+DAS & 0.17 & 0.26 & 0.08 & 0.15 \\
  & S2+DAS & \textbf{0.63} & \textbf{0.75} & \textbf{0.70} & \textbf{0.82} \\
  & S1+S2+DAS & 0.61 & 0.74 & 0.63 & 0.78 \\
\addlinespace
3 & S1+DAS & 0.33 & 0.45 & 0.23 & 0.37 \\
  & S2+DAS & \textbf{0.77} & \textbf{0.87} & \textbf{0.77} & \textbf{0.87} \\
  & S1+S2+DAS & 0.70 & 0.81 & 0.70 & 0.82 \\
\addlinespace
4 & S1+DAS & 0.72 & 0.82 & 0.72 & 0.82 \\
  & S2+DAS & \textbf{0.87} & \textbf{0.92} & \textbf{0.87} & \textbf{0.92} \\
  & S1+S2+DAS & 0.84 & 0.91 & 0.84 & 0.91 \\
\bottomrule
\end{tabular}
\label{tab:per_split_results_ssl}
\end{table*}

\begin{table*}[!htbp]
\centering
\setlength{\tabcolsep}{3pt}
\renewcommand{\arraystretch}{1.1}
\caption{Aggregate performance of AnySat encoders trained from scratch across 4 cross-validation splits (images with $>$5\% water only).}
\begin{tabular}{@{}l c c c c@{}}
\toprule
\textbf{Modalities} & \textbf{Mean IoU} & \textbf{Mean F1} & \textbf{Median IoU} & \textbf{Median F1} \\
\midrule
  S1+DAS (from scratch) & $0.48 \pm 0.30$ & $0.58 \pm 0.30$ & $0.50 \pm 0.35$ & $0.60 \pm 0.35$ \\
  S2+DAS (from scratch) & $\mathbf{0.71 \pm 0.15}$ & $\mathbf{0.81 \pm 0.12}$ & $\mathbf{0.74 \pm 0.14}$ & $\mathbf{0.84 \pm 0.09}$ \\
  S1+S2+DAS (from scratch) & $0.70 \pm 0.15$ & $0.81 \pm 0.12$ & $0.72 \pm 0.13$ & $0.83 \pm 0.09$ \\
\bottomrule
\end{tabular}
\label{tab:aggregate_results_scratch}
\end{table*}

\begin{table*}[!htbp]
\centering
\footnotesize
\renewcommand{\arraystretch}{1.1}
\caption{Per-split performance of AnySat encoders trained from scratch across validation sets, restricted to images with $>$5\% water.}
\begin{tabular}{@{}c l c c c c@{}}
\toprule
\textbf{Split} & \textbf{Modalities} & \textbf{Mean IoU} & \textbf{Mean F1} & \textbf{Median IoU} & \textbf{Median F1} \\
\midrule
1 & S1+DAS (from scratch) & 0.54 & 0.68 & 0.55 & 0.71 \\
  & S2+DAS (from scratch) & \textbf{0.67} & \textbf{0.79} & \textbf{0.68} & \textbf{0.81} \\
  & S1+S2+DAS (from scratch) & 0.66 & 0.78 & 0.65 & 0.79 \\
\addlinespace
2 & S1+DAS (from scratch) & 0.08 & 0.13 & 0.03 & 0.06 \\
  & S2+DAS (from scratch) & \textbf{0.52} & \textbf{0.66} & \textbf{0.57} & \textbf{0.73} \\
  & S1+S2+DAS (from scratch) & 0.52 & 0.66 & 0.57 & 0.73 \\
\addlinespace
3 & S1+DAS (from scratch) & 0.50 & 0.65 & 0.56 & 0.72 \\
  & S2+DAS (from scratch) & 0.75 & 0.85 & 0.80 & 0.89 \\
  & S1+S2+DAS (from scratch) & \textbf{0.77} & \textbf{0.87} & \textbf{0.81} & \textbf{0.89} \\
\addlinespace
4 & S1+DAS (from scratch) & 0.79 & 0.87 & 0.85 & 0.92 \\
  & S2+DAS (from scratch) & \textbf{0.88} & \textbf{0.94} & \textbf{0.89} & \textbf{0.94} \\
  & S1+S2+DAS (from scratch) & 0.87 & 0.93 & 0.86 & 0.92 \\
\bottomrule
\end{tabular}
\label{tab:per_split_results_scratch}
\end{table*}

\subsection{Damage Assessment Application}
The 2019 flood in Tulun, Russia, was selected as a case study to demonstrate the practical application of the proposed damage assessment framework. The analysis focuses on Tulun and the adjacent upstream area along the Iya River, bounded by the geographic coordinates 54.649008°N, 100.440445°E and 54.215066°N, 100.828571°E and  in the WGS84 coordinate system.

\begin{figure*}[htbp]
    \centering
    \includegraphics[width=1\textwidth]{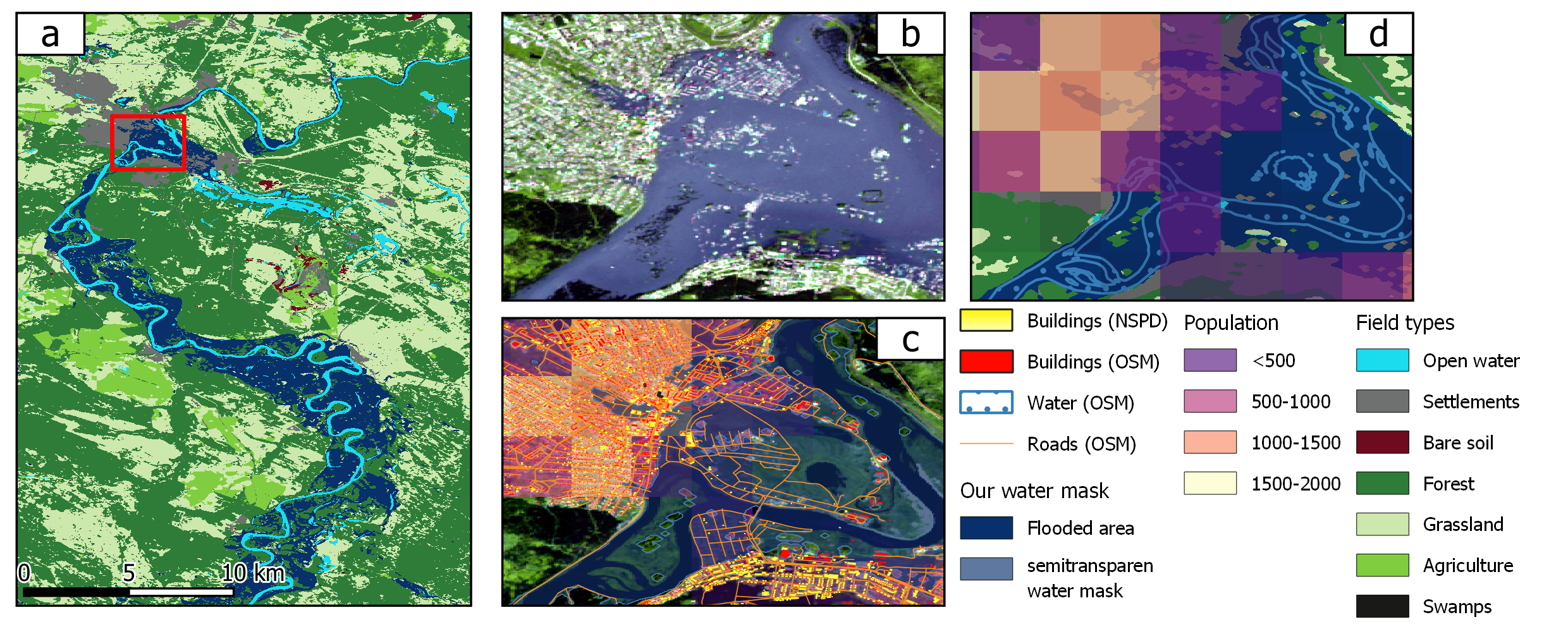}
    \caption{Visualization of flood damage assessment in the Tulun region. 
    \textbf{(a)} Overview of the damage assessment case study area, showing the flood water mask overlaid on a land cover classification map. The red box indicates the location of panels~(b)--(d). 
    \textbf{(b)} S2 satellite image from 29 June 2019, depicting the inundation of Tulun city. 
    \textbf{(c)} Data from NSPD and OSM: infrastructure, road network; and population density map, overlaid with a semi-transparent flood mask on a pre-flood S2 image (19 June 2019).
    \textbf{(d)} Flood mask with river channel delineation (OSM), land cover classification, and population density map for Tulun.}
    \label{fig:damage_ass}
\end{figure*}

Figure~\ref{fig:damage_ass} illustrates the flood damage assessment workflow. Panel (a) shows the entire case study area overlaid with the flood water mask and land cover classifications, enabling rapid identification of high-risk zones. Panel (b) displays a S2 satellite image captured during the 2019 flood event. Panels (c) and (d) overlay population distribution and infrastructure data (including buildings and roads) for Tulun onto a pre-flood S2 image, revealing that a significant portion of the city’s populated area—along with associated buildings and road networks—lies within the inundated zone.
A key methodological limitation arises from the timeliness of the NSPD and OSM datasets. Both databases are updated regularly; consequently, structures destroyed during the 2019 flood and subsequently removed from official or crowdsourced records are no longer represented in these sources. As a result, the methodology cannot capture such losses. This limitation is clearly illustrated in Figure~\ref{fig:damage_ass} b and c, where S2 imagery reveals buildings that are absent from current NSPD and OSM data. The framework is therefore inherently dependent on the completeness and temporal alignment of its input datasets, NSPD data exhibiting notable gaps in this case.
The analysis of the damage revealed the following parameters: a total inundated area of 112.78 km², and 14.67 km² of them fell within the administrative boundaries of Tulun city. The permanent water bodies area mapped in OpenStreetMap is 27.48 km². This estimate for Tulun shows strong agreement with official reports, which indicate a maximum flooded area of approximately 13–16 km²~\cite{roscosmos2023, irkcity2019}.

For the analysis of the 2019 Tulun flood, the 2020 population data with 30 arcseconds resolution from GHS-POP dataset was used as the closest available approximation. The analysis estimated that approximately 15\% of the population of the study area falls
within the modeled inundation extent. Official casualty statistics reports 26~\cite{tass2020} fatalities across the entire affected region. Evacuated people total amount is more than two thousand~\cite{tass2019}. Current approach doesn't consider all EMERCOM evacuation protocols, that's why the direct comparison between the estimates and official reports is not possible. 
Building exposure analysis based on NSPD cadastral data identified 162 structures within the flood extent, including 101 residential buildings and 149 were classified as privately owned. The official post-disaster report, which stated that 83 buildings had been completely destroyed by floods in Tulun. However, OSM data indicates that the number of buildings potentially affected is 2,178. This significant discrepancy between two geospatial datasets highlights a problem that is relevant for Russia and elsewhere: the presence of illegal or undeclared buildings that cannot be accounted with the government. According to official statistics, 2,451 buildings were damaged as a result of the flooding \cite{irk_ru_2024_tulun_flood}.
The mismatch between cadastral and actual statistics leads to discrepancies in economic damage assessments. The aggregate cadastral value of buildings within the flooded area is estimated at approximately 500 million RUB. This figure contrasts sharply with official damage assessments for residential properties, which reported losses of approximately 17~\cite{irkcity2019} billion RUB. The approximate damage estimate, based on the number of damaged buildings according to OSM data and the total cadastral value, is approximately 7 billion rubles, which is a fairly accurate result. This substantial discrepancy can be attributed to two primary factors: (1) the methodology’s reliance on cadastral values-which typically reflect market or tax-assessed values, rather than replacement or reconstruction costs—and (2) the incompleteness of the underlying building datasets, as previously discussed.

In terms of ecological and agricultural impact, the model indicates that the most extensively flooded land cover types were grasslands (62 km²) and forests (37 km²), whereas only 0.9 km² of agricultural land was inundated. The least affected category was bare soil, with approximately 0.09 km² flooded. The inundated area overlaps with habitats of 1,342 plant taxa and 23 mammal taxa. 
According to the methodology, 25 bird taxa listed in the Red Data Book of the Irkutsk Region are recorded in the study area. However, given the high mobility of avian species and their limited dependence on ground-level habitats during flood events, it is unlikely that these taxa will experience direct harm from flooding.

The proposed workflow provides a fully automated, multimodal system for rapid flood impact assessment. It visualizes inundated areas with key geospatial data: population density, infrastructure, agricultural land, and biodiversity. 

The framework is highly adaptive, and morphological closure and buffering parameters can be optimised for specific tasks. For example, field types area calculation requires minimal buffering of water mask, while evacuations planning benefit from an expanded buffer zone. Furthermore, indirect water level estimations are possible through the workflow. By correlating flooded buildings with cadastral data on the number of floors, the system can infer flood depths, providing valuable validation where direct hydrological measurements are unavailable.

\section{Conclusion}

In this study we introduced and evaluated a multimodal flood monitoring framework adapted for working in Russia, leveraging radar, multispectral, and digital elevation model data. The framework is designed for scenarios with limited labelled data, following the operational requirements of the Russian Ministry of Emergency Situations, and comprises two main components: water surface detection and flood damage assessment.

Since the water surface detection task involves working with small amounts of multimodal data, two approaches were considered and compared: supervised and self-supervised based. For supervised water surface detection Unet++ architecture was selected. As other studies have demonstrated the great potential of the self-learning approach for working with multimodal satellite data, it was decided to test its applicability to this task. The AnySat architecture was chosen because it allows work with any configuration of modalities. The following combinations of satellite data were considered: radar (S1), multispectral (S2), and digital elevation model (DEM). The following experiments were conducted: training the UNet++ architecture for water surface detection, pre-training AnySat using the SSL approach, followed by subsequent retraining for the segmentation task, direct training of the AnySat architecture from scratch to solve the water surface segmentation task. While SSL pre-training improved fold-to-fold stability across all modality combinations, the U-Net++ baseline achieved the highest overall segmentation quality, with a mean IoU of $0.75 \pm 0.15$ and a mean F1-score of $0.84 \pm 0.11$. The limited size of the SSL pre-training corpus, GPU-memory-driven input downscaling, and suboptimal cross-modal feature alignment are likely the main factors that prevented the SSL pathway from realising the conceptual advantages of foundation-style models in this study. Therefore, under the data and task conditions targeted here supervised learning on labelled data remains the most effective approach, while the AnySat-based pathway retains its appeal for settings in which larger unlabelled corpora, robustness to missing modalities, or backbone reuse across multiple downstream tasks become operationally important.

The model predictions that performed best were used to create a flood damage assessment framework capable of automatically assessing the extent of inundation, population exposure and building-level economic losses. The flood in Tulun in 2019 was considered as a case study. The proposed workflow produced results that closely matched official post-event reports, confirming the applicability of the framework for rapid, data-driven disaster assessment. By combining multimodal satellite imagery with cadastral and population data, the framework can quantify the physical, economic and ecological impacts of flooding and infer approximate flood depths based on structural characteristics.

Several challenges remain. The main limitations identified in this study are the computational cost of multimodal pre-training, the scarcity of temporally aligned datasets and the limited revisit frequency of satellite imagery. Future work should focus on expanding the pre-training corpus, integrating additional modalities such as thermal infrared or LiDAR, and using federated learning to incorporate regional data while keeping sensitive information decentralised. Furthermore, integrating this framework with near-real-time satellite acquisition and hydrological forecasting systems would facilitate proactive flood risk prediction instead of post-event assessment.

\section*{Statements \& Declarations}

\subsection*{Competing Interests}
The authors have no relevant financial or non-financial interests to disclose.

\subsection*{Data Availability}
The datasets used and analysed during the current study available from the corresponding author on reasonable request.

\subsection*{Author Contributions}

I.N., R.D., M.S., A.K., and S.I. conducted the conceptualization, I.N., R.D., and M.S. designed the methodology, I.N., R.D., M.S., and A.A. developed the software, I.N., R.D., and M.S. performed the validation, I.N., R.D., M.S., and A.A. conducted the investigation, I.N. and R.D. performed data curation, I.N., R.D., M.S., A.A., and M.U. contributed to visualization, S.I., D.S., and E.B. supervised the project, I.N. and S.I. administered the project, I.N. prepared the original draft, I.N., R.D., M.S., A.A., M.U., and S.I. reviewed and edited the manuscript. 

\backmatter

\begin{appendices}

\section{Qualitative Results}\label{secA1}

We visualize representative validation scenes to illustrate typical success and failure modes under three input configurations:
\emph{S1 + DAS}, \emph{S2 + DAS}, and \emph{S1 + S2 + DAS}, where
\textbf{S1} = Sentinel\mbox{-}1 SAR backscatter,
\textbf{S2} = Sentinel\mbox{-}2 multispectral,
and \textbf{DAS} = DEM with topographic derivatives (aspect, slope).

\begin{figure*}[htbp]
\centering
\includegraphics[width=0.8\linewidth]{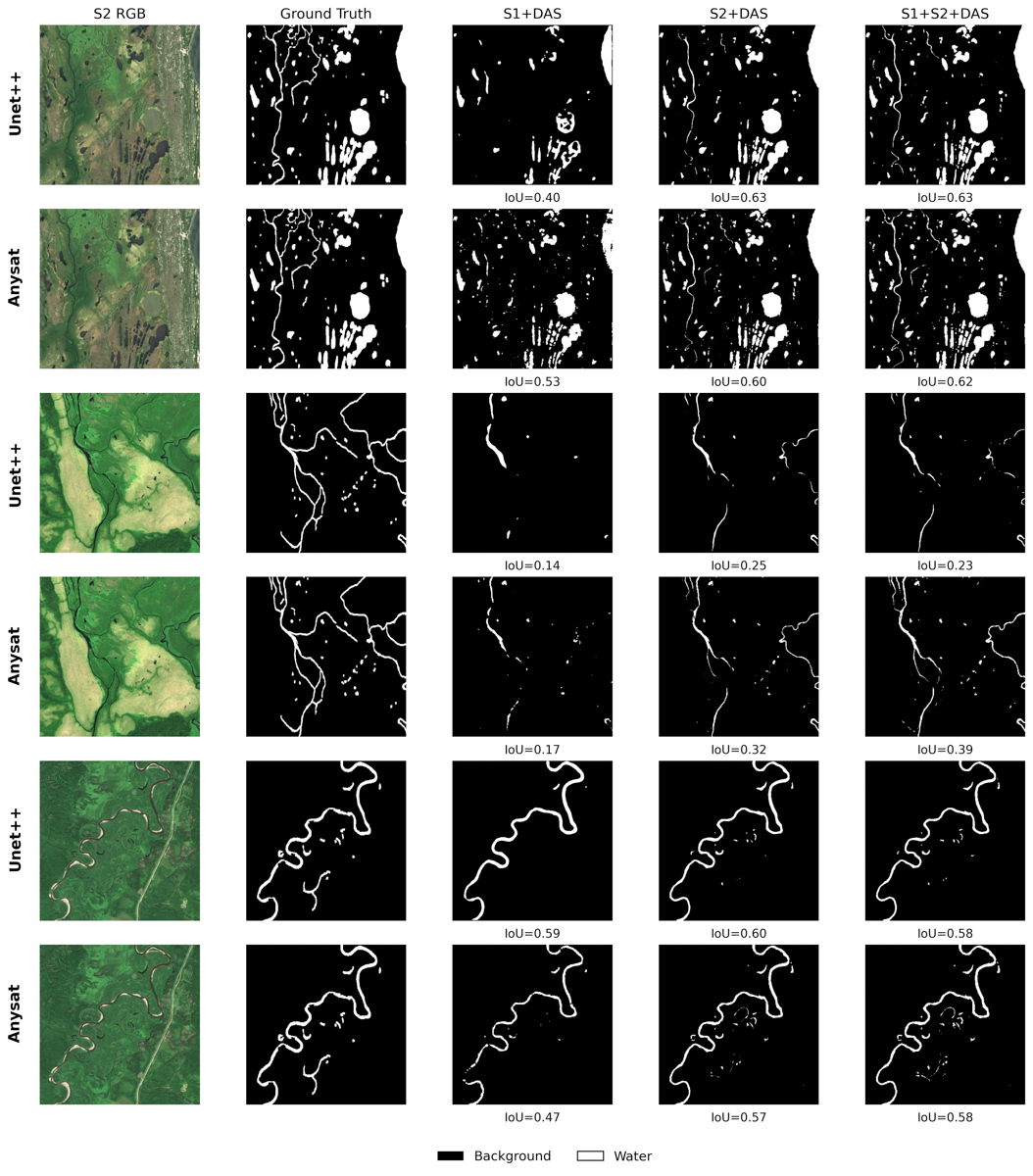}
\caption{
Qualitative examples of the flood-extent segmentation with the Unet++ and AnySat-based (trained from scratch) models under three input configurations for validation split \#1. Columns from left to right: S2 input image (RGB channels), ground-truth mask, model prediction with different inputs: S1 + DAS, S2 + DAS, and S1 + S2 + DAS.
}
\label{fig:models_results_split1}
\end{figure*}

\begin{figure*}[htbp]
\centering
\includegraphics[width=0.8\linewidth]{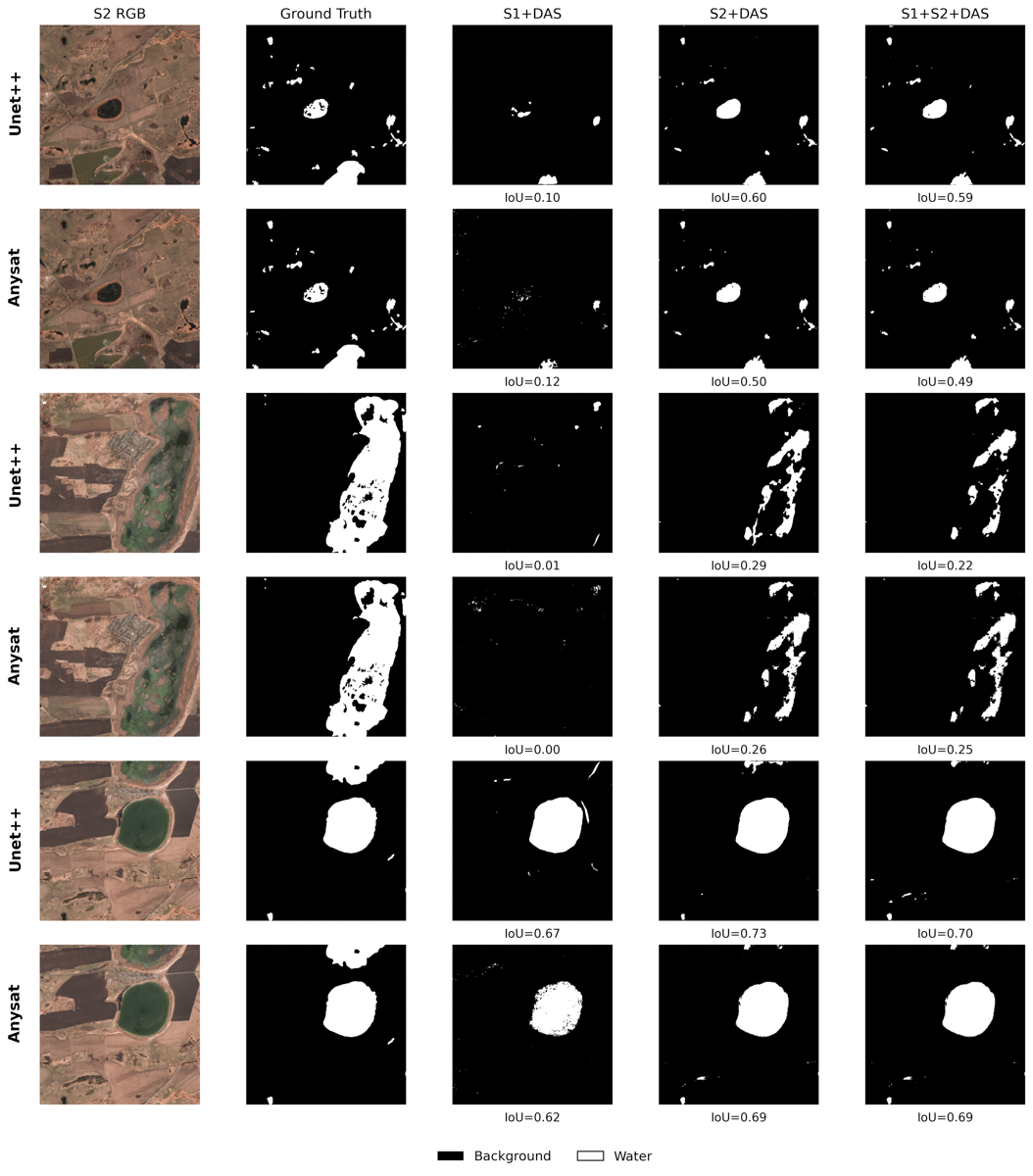}
\caption{
Qualitative examples of the flood-extent segmentation with the Unet++ and AnySat-based (trained from scratch) models under three input configurations for validation split \#2. Columns from left to right: S2 input image (RGB channels), ground-truth mask, model prediction with different inputs: S1 + DAS, S2 + DAS, and S1 + S2 + DAS.
}
\label{fig:models_results_split2}
\end{figure*}

\begin{figure*}[htbp]
\centering
\includegraphics[width=0.8\linewidth]{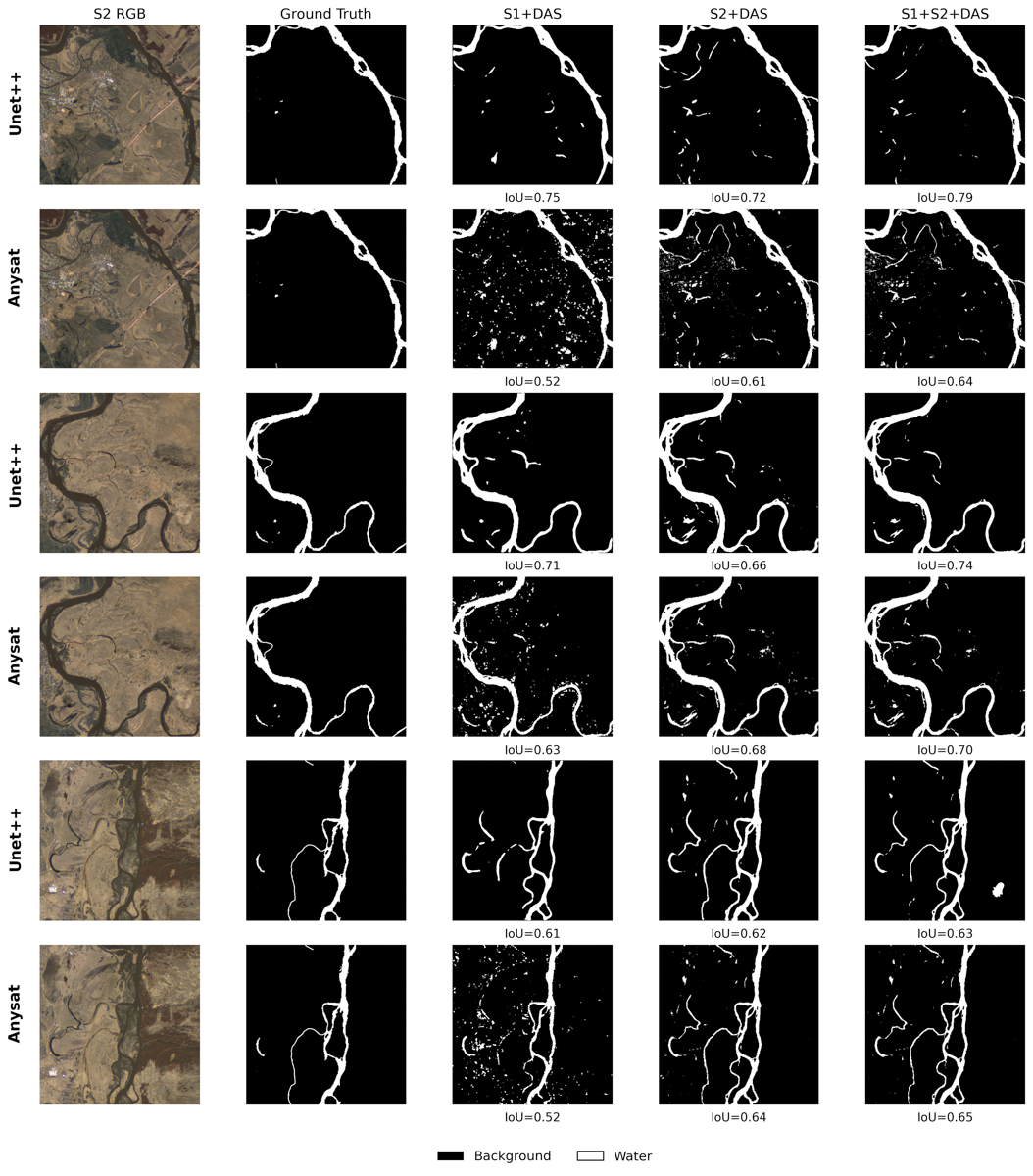}
\caption{
Qualitative examples of the flood-extent segmentation with the Unet++ and AnySat-based (trained from scratch) models under three input configurations for validation split \#3. Columns from left to right: S2 input image (RGB channels), ground-truth mask, model prediction with different inputs: S1 + DAS, S2 + DAS, and S1 + S2 + DAS.
}
\label{fig:models_results_split3}
\end{figure*}

\begin{figure*}[htbp]
\centering
\includegraphics[width=0.8\linewidth]{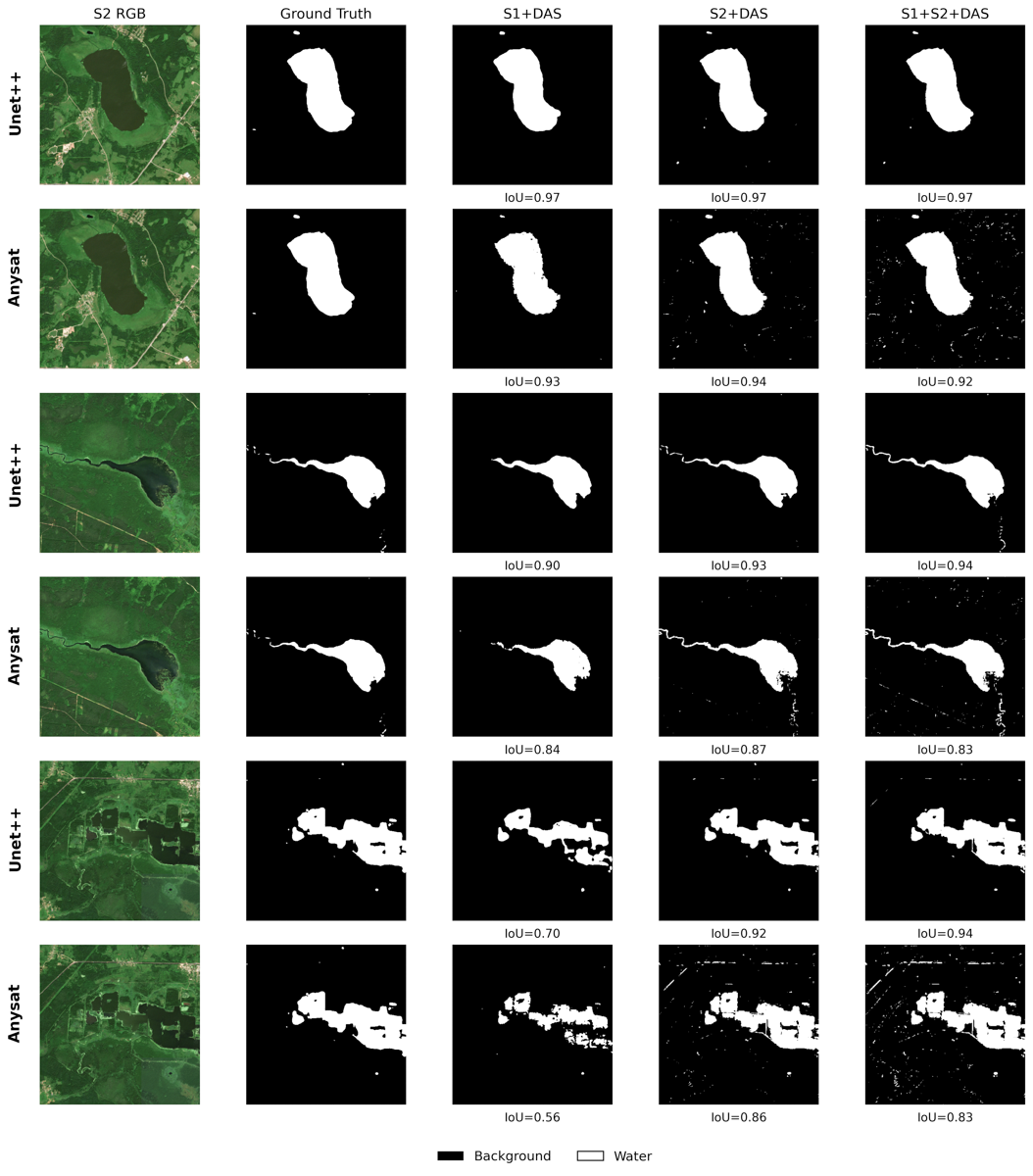}
\caption{
Qualitative examples of the flood-extent segmentation with the Unet++ and AnySat-based (trained from scratch) models under three input configurations for validation split \#4. Columns from left to right: S2 input image (RGB channels), ground-truth mask, model prediction with different inputs: S1 + DAS, S2 + DAS, and S1 + S2 + DAS.
}
\label{fig:models_results_split4}
\end{figure*}

\bibliography{sn-bibliography}

\end{appendices}
\end{document}